%% file: main.tex
\documentclass[conference]{IEEEtran}
\usepackage{cite}
\usepackage{amsmath,amssymb,amsfonts}
\usepackage{algorithmic}
\usepackage{graphicx}
\usepackage{textcomp}
\usepackage{xcolor}
\usepackage{fancyhdr}
\usepackage[hyphens]{url}
\usepackage{soul}
\usepackage[utf8]{inputenc} 
\usepackage[T1]{fontenc}    
\usepackage{hyperref}       
\usepackage{url}            
\usepackage{booktabs}       
\usepackage{amsfonts}       
\usepackage{nicefrac}       
\usepackage{microtype}      
\usepackage[export]{adjustbox}
\usepackage{caption}
\usepackage{subcaption}
\usepackage{xcolor}
\usepackage{xspace}
\usepackage{multirow}
\usepackage{bbold}
\usepackage{lipsum}
\usepackage{amssymb}
\usepackage{pifont}
\usepackage{listings} 
\newcommand{\cmark}{\ding{51}}%
\usepackage{amsmath} 
\usepackage{float}
\usepackage{caption}
\usepackage{pbox}

\usepackage[normalem]{ulem}
\usepackage{listings}

\def\BibTeX{{\rm B\kern-.05em{\sc i\kern-.025em b}\kern-.08em
    T\kern-.1667em\lower.7ex\hbox{E}\kern-.125emX}}

\definecolor{dkgreen}{rgb}{0,0.6,0}
\definecolor{gray}{rgb}{0.5,0.5,0.5}
\definecolor{mauve}{rgb}{0.58,0,0.82}

\lstdefinestyle{top}{
  float=t,
  floatplacement=t,
  abovecaptionskip=2pt
}

\restylefloat{table} 

\newif\ifcomments
\commentstrue
\commentsfalse

\ifcomments
\usepackage[textsize=small,color=lightgray]{todonotes}
\setuptodonotes{fancyline}
\else
\usepackage[disable]{todonotes}
\fi

\newcommand{\jenny}[1]{\todo[ color=cyan!50]{\textbf{Jenny}: #1}\ignorespaces}

\newcommand{\joshil}[1]{\todo[inline, color=LimeGreen!50]{\textbf{Josh}: #1}\ignorespaces}

\renewcommand*{\figureautorefname}{Fig.}

\newcommand{\asplossubmissionnumber}{378}

\newcommand{\sys}[0]{CoSA\xspace}

\title{\sys: Scheduling by \underline{C}onstrained \underline{O}ptimization for \underline{S}patial \underline{A}ccelerators} 

\makeatletter
\newcommand{\linebreakand}{%
  \end{@IEEEauthorhalign}
  \hfill\mbox{}\par
  \mbox{}\hfill\begin{@IEEEauthorhalign}
}
\makeatother

\begin{document}
\setlength{\parskip}{0.2mm}
\setlength{\dblfloatsep}{0.2cm}
\setlength{\dbltextfloatsep}{0.2cm}
\setlength{\floatsep}{0.2cm}
\setlength{\textfloatsep}{0.2cm}

\author{
\IEEEauthorblockN{Qijing Huang}
\IEEEauthorblockA{\textit{UC Berkeley}}
\and 
\IEEEauthorblockN{Minwoo Kang}
\IEEEauthorblockA{\textit{UC Berkeley}}
\and
\IEEEauthorblockN{Grace Dinh}
\IEEEauthorblockA{\textit{UC Berkeley}}
\and
\IEEEauthorblockN{Thomas Norell}
\IEEEauthorblockA{\textit{UC Berkeley}} 
\linebreakand
\IEEEauthorblockN{Aravind Kalaiah}
\IEEEauthorblockA{\textit{Facebook}}
\and 
\IEEEauthorblockN{James Demmel}
\IEEEauthorblockA{\textit{UC Berkeley}}
\and
\IEEEauthorblockN{John Wawrzynek}
\IEEEauthorblockA{\textit{UC Berkeley}}
\and
\IEEEauthorblockN{Yakun Sophia Shao}
\IEEEauthorblockA{\textit{UC Berkeley}}
}

\maketitle

\pagestyle{plain}

\input{0abstract}

\input{1intro}

\input{2background}

\input{3framework}
\input{4methodology}
\input{5evalution}
\input{6conclusion}
\input{8ack}
\clearpage
\bibliographystyle{IEEEtranS}
\bibliography{references}




\end{document}

%% file: 0abstract.tex
\begin{abstract}

Recent advances in Deep Neural Networks (DNNs) have led to active development of
specialized DNN accelerators, many of which feature a large number of processing
elements laid out spatially, together with a multi-level memory hierarchy and
flexible interconnect.
While DNN accelerators can take advantage of data reuse and achieve high peak throughput, they also expose a large number of runtime parameters to the
programmers who need to explicitly manage how computation is
scheduled both \textit{spatially} and \textit{temporally}.
In fact, different scheduling choices can lead to wide variations in performance
and efficiency, motivating the need for a fast and efficient
search strategy to navigate the vast scheduling space.

To address this challenge, we present \sys, a constrained-optimization-based approach
for scheduling DNN accelerators.
As opposed to existing approaches that either rely on designers' heuristics or iterative methods to navigate 
the search space, \sys expresses scheduling decisions as a constrained-optimization problem
that can be deterministically solved using mathematical optimization techniques.
Specifically, \sys leverages the regularities in DNN operators and hardware to
formulate the DNN scheduling space into a mixed-integer programming (MIP) problem with
algorithmic and architectural constraints, which can be solved to automatically generate
a highly efficient schedule in one shot. 
We demonstrate that \sys-generated schedules significantly outperform
state-of-the-art approaches by a geometric mean of up to 2.5$\times$ across a wide range of DNN
networks while improving the time-to-solution by 90$\times$.
\end{abstract}

%% file: 1intro.tex
\section{Introduction}
\label{sec:intro}

Deep neural networks 
(DNNs) have gained major interest in
recent years due to their robust ability to learn based on large amounts of data.  
DNN-based approaches have been applied to computer
vision~\cite{alexnet, resnet, yolo}, machine
translation~\cite{sutskever2014sequence, transformer}, audio
synthesis~\cite{wavenet}, recommendation models~\cite{naumov2019dlrm,
gupta-dlrm-hpca2020},
autonomous driving~\cite{drivenet} and many other fields.  Motivated by the high
computational requirements of DNNs, there have been exciting developments in both
research and commercial spaces in building specialized DNN accelerators for both
edge\cite{eyeriss-isca2016, diannao, shidiannao-isca2015, cambricon, scnn, tpu_edge,
nvdla-hotchips, samsung} 
and cloud applications~\cite{dadiannao, scaledeep, tpu-isca2016, brainwave-isca-2018,
aws-inferentia, centaur-isca2020}.

State-of-the-art DNN accelerators typically incorporate large arrays of
processing elements to boost parallelism, together with a deep multi-level memory
hierarchy and a flexible network-on-chip (NoC) to improve data reuse.
While these architectural structures can improve the performance and
energy efficiency of DNN execution, they also expose a large number of
scheduling parameters to programmers who must decide when and where each piece of computation and data movement is mapped onto the accelerators both
spatially and temporally.
Here, we use \textit{schedule} to describe how a DNN layer is partitioned
spatially and temporally to execute on specialized accelerators.
Given a target DNN layer and a specific hardware architecture, there
could be millions, or even billions, of valid schedules with a wide range of performance and energy efficiency~\cite{timeloop2019-ispass}.
Considering the vast range of DNN layer dimensions and hardware
architectures, there is a significant demand for a generalized framework to
quickly produce efficient scheduling options for accelerators of
varying hardware configurations. 

Achieving high performance on a spatially distributed architecture requires several factors
to be carefully considered, including tiling for good hardware utilization,
pipelining data movement with compute,
and maximizing data
re-use.  Previous scheduling frameworks have attempted to reflect these
considerations by formulating an analytical cost model, pruning the scheduling space
with known hardware constraints, and then exhaustively searching for the best
candidate based on their cost models~\cite{timeloop2019-ispass, interstellar-asplos2020, chatarasi2020marvel, dave2019dmazerunner}.  However,
navigating the scheduling space in such a brute-force fashion can easily become
intractable for larger DNN layers and more complex hardware architectures.
Other notable efforts have employed feedback-driven approaches, such as black-box tuning, beam search, and other machine learning algorithms with iterative
sampling~\cite{tvm2018-osdi, jia2019beyond, adams2019learning}.
However, these schedulers typically require massive
training datasets and large-scale simulations to learn performance models,
making it infeasible to extend them to other types of hardware accelerators, especially
those still under development.
Hence, there is a clear need for efficient scheduling mechanisms to
\textit{quickly} navigate the search space and produce \textit{performant}
scheduling options.

In this work, we demonstrate \sys{}, a constrained-optimization-based approach to schedule DNN accelerators.
In contrast to prior work that either requires exhaustive brute-force-based or expensive
feedback-driven approaches, \sys expresses the DNN accelerator scheduling as a
constrained-optimization problem that can be deterministically solved using
today's mathematical optimization libraries in one pass.
In particular, \sys leverages the regularities in both DNN layers and spatial
hardware accelerators where the algorithmic and hardware parameters can be
clearly defined as scheduling constraints.
Specifically, \sys formulates the DNN scheduling problem as a prime-factor
allocation problem that determines 1) tiling sizes for different memory levels, 2) relative loop ordering to exploit reuse, and 3) how computation should be executed spatially and temporally.
\sys constructs the scheduling constraints by exposing both the
algorithmic behaviors, e.g., layer dimensions, and hardware parameters, e.g., memory and network hierarchies.
Together with clearly defined and composable objective functions, \sys can solve the
DNN scheduling problem in one shot without expensive iterative
search.
Our evaluation demonstrates that \sys-generated schedules outperform
state-of-the-art approaches by $2.5\times$ across different DNN network
layers, while requiring $90\times$ less scheduling time as it does not require iterative search. 

\begin{figure}[t]
  \centering
  \includegraphics[width=0.98\linewidth]{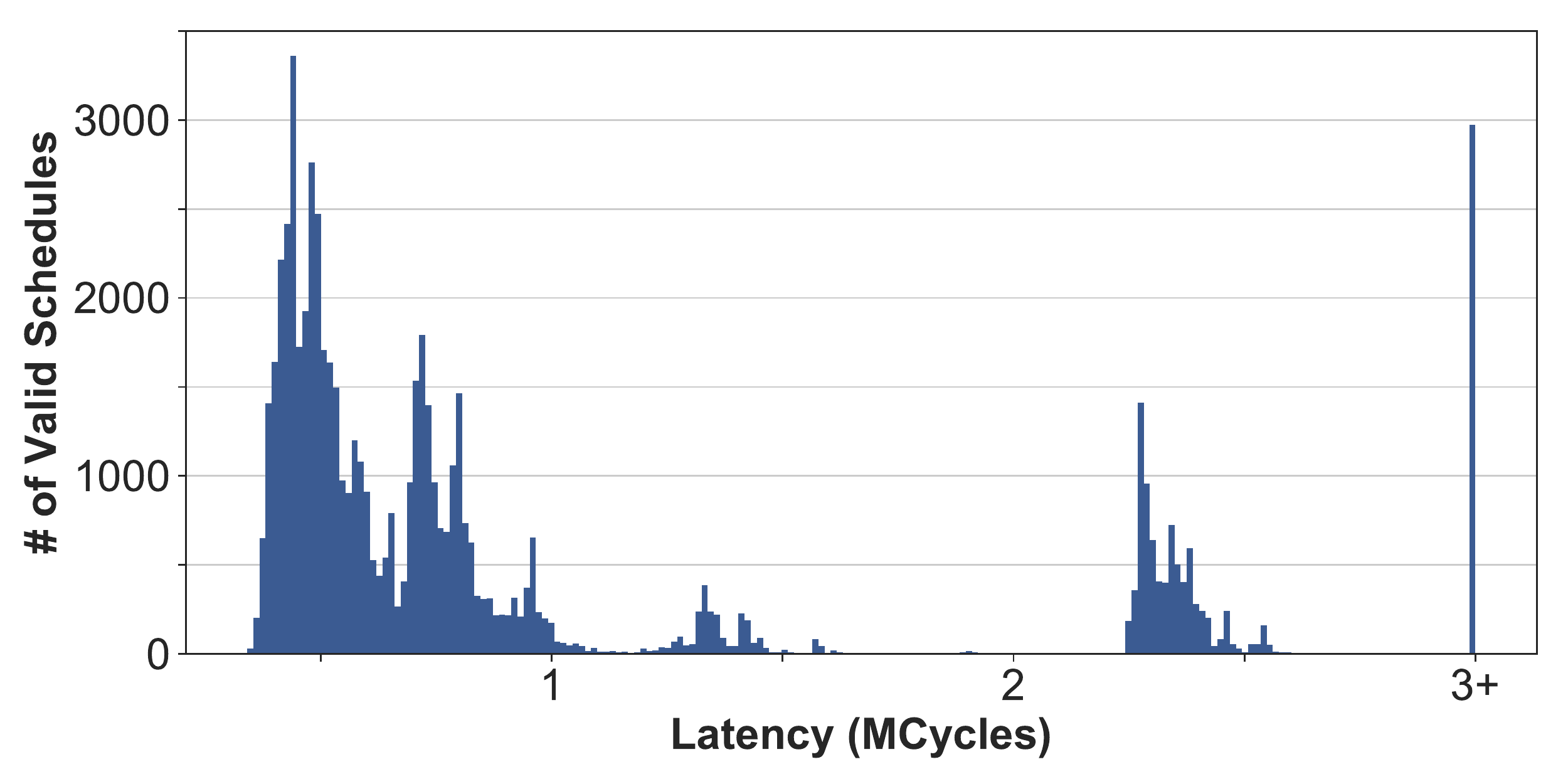}
  \vspace{-5pt}
  \caption{\small Execution latency histogram of 40K valid scheduling choices for
  a ResNet-50 layer on a spatial accelerator.}
  \label{fig:tilling_characterization}
\end{figure} 
In summary, this work makes the following contributions:
\begin{itemize}
    \item We formulate DNN accelerator scheduling as a constrained-optimization problem
     that can be solved in a single pass.
    To the best of our knowledge, \sys is the first constrained-optimization-based
    approach to tackle major DNN scheduling decisions in one shot. 

    \item We take a communication-oriented approach in the \sys formulation that
    highlights the importance of data transfer across different on-chip
    memories and exposes the cost through clearly defined objective functions.

    \item We demonstrate that \sys can quickly generate high-performance schedules
    outperforming state-of-the-art approaches for different DNN layers
    across different hardware architectures.

\end{itemize}

%% file: 2background.tex
\section{Background and Motivation}
\label{sec:background}
In this section, we discuss the complexity of DNN scheduling space and the state-of-the-art schedulers to navigate the space.

\subsection{DNN Scheduling Space}
\label{sec:bg_motivation}

Scheduling is a crucial decision-making process for the compilers to effectively assign workload to compute resources. 
With the emergence of numerous DNN accelerators with diverse architectures, 
there is a need for a fast, performant, and
explainable approach to scheduling. 
Our work focuses on operator-level scheduling, which aims to optimize the performance of each operator, i.e. DNN layer, on specific hardware. 
Operator-level
scheduling typically comprises three key loop optimizations: \textit{loop
tiling}, \textit{loop permutation}, and \textit{spatial mapping}.  
\textit{Loop tiling} describes which loops are mapped to which memory hierarchy and the corresponding tile sizes.
\textit{Loop permutation}
determines the relative order of the loops, while \textit{spatial mapping} binds
one or more loop dimensions to spatial hardware resources, such as parallel processing elements, instead of mapping them to temporal (i.e. sequential) execution. Each optimization can have a significant impact on the performance, and all three optimizations need to be considered together to achieve the best performance.  



Consider scheduling a 3$\times$3 convolution layer in ResNet50~\cite{resnet} 
with 256 input and output channels, and an output dimension of 14$\times$14, 
on an accelerator with five levels of memory.
If we split each individual loop bound into its prime factors and assign each one to a memory level, we would have billions of schedules to consider.
Among the randomly sampled schedules from all
possible loop tilings, half of them fail to satisfy the buffer capacity constraints (e.g. a schedule is invalid if it requires a 4KB buffer, though the available buffer size is only 2KB.). 
\figureautorefname{}~\ref{fig:tilling_characterization} shows the performance distribution of the valid schedules.
We observe a wide performance difference among the valid schedules, with the best one outperforming the worst one by $7.2\times$.
In addition, we observe clusters of schedules that have similar latencies in the \figureautorefname{}~\ref{fig:tilling_characterization}, revealing structure in the solution space.

\subsection{State-of-the-art Schedulers}



\begin{table}[t]
\footnotesize
\adjustbox{width=\linewidth}{
\begin{tabular}{cc}
\toprule
Scheduler &
Search Algorithm \\
\midrule
\midrule

\multicolumn{2}{l}{\textit{Brute-force Approaches:} } \vspace{3pt}\\
Timeloop~\cite{timeloop2019-ispass} & Brute-force \& Random  \\
dMazeRunner~\cite{dave2019dmazerunner} & Brute-force  \\
Triton~\cite{tillet2019triton} &{Brute-force over powers of two}\\
Interstellar~\cite{interstellar-asplos2020} & Brute-force  \\
Marvel~\cite{chatarasi2020marvel} & Decoupled  Brute-force  \\
\midrule
\multicolumn{2}{l}{\textit{Feedback-based Approaches:} } \vspace{3pt}\\
AutoTVM ~\cite{tvm2018-osdi} & ML-based Iteration \\
Halide~\cite{ragan2013halide} & Beamsearch~\cite{adams2019learning},
OpenTuner~\cite{ansel2014opentuner, mullapudi2016automatically} \\
FlexFlow~\cite{jia2019beyond} & MCMC \\
Gamma~\cite{gamma-iccad2020} & Genetic Algorithm \\
Mind Mapping~\cite{hegde2021mind} & Gradient-based Search \\
\midrule
\multicolumn{2}{l}{\textit{Constrained Optimization Approaches:} } \vspace{3pt}\\
Polly+Pluto~\cite{grosser2011polly, bondhugula2008practical, bondhugula2016pluto+} &  \\
Tensor Comprehension~\cite{vasilache2018tensor} &  Polyhedral Transformations \\ 
Tiramisu~\cite{bagehadi2019tiramisu} &  \\
\midrule
\textbf{CoSA} & \bf{Mixed-Integer Programming (MIP)} \\
\bottomrule
\end{tabular}
\caption{\small State-of-the-art DNN accelerator schedulers.\label{table:related_work}}
}
\end{table}
\smallskip  

\label{sec:framework} 
\begin{figure*}[t]
   \begin{minipage}{0.67\textwidth}
      \centering
      \includegraphics[width=0.95\linewidth]{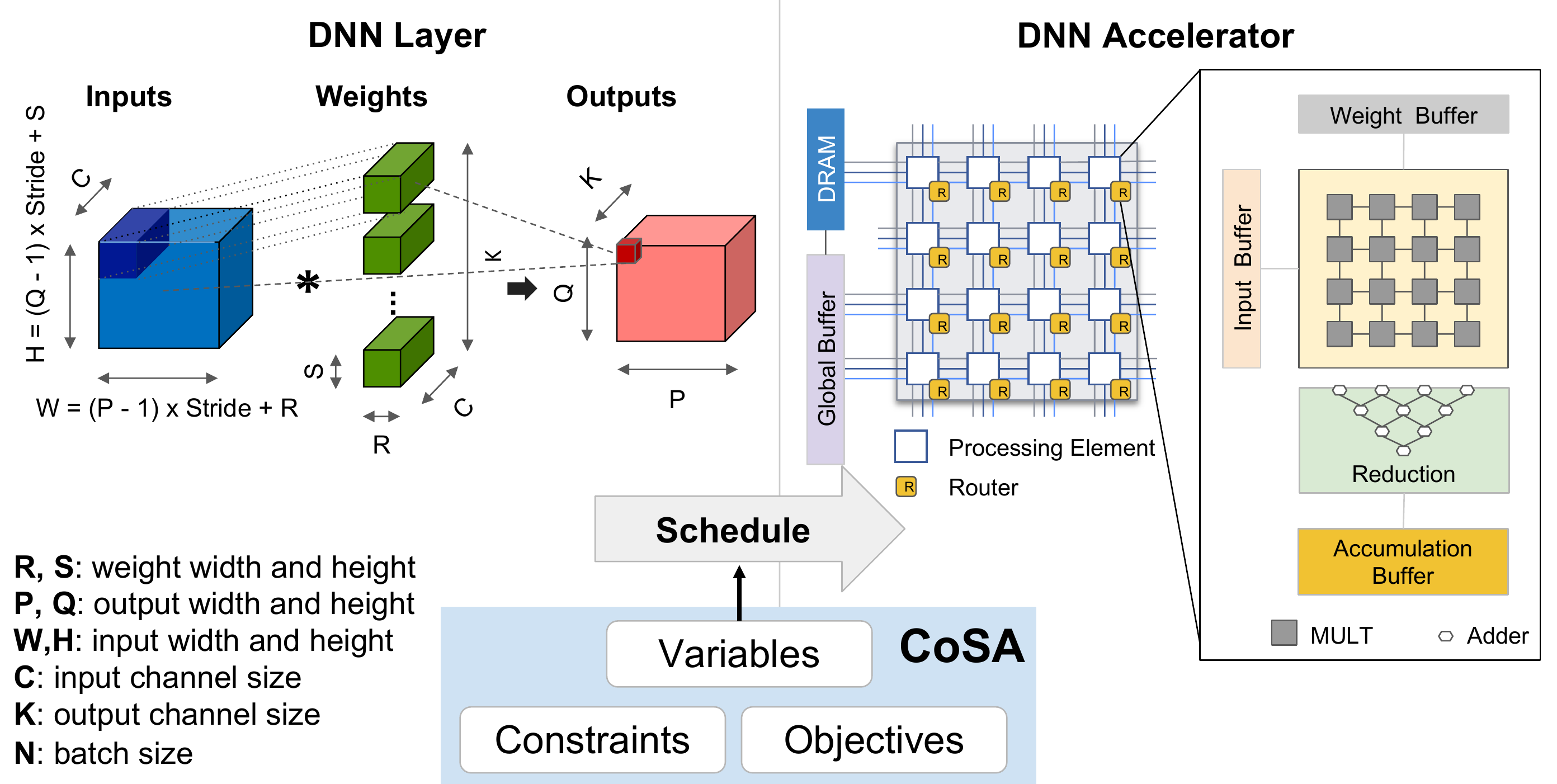}
      \caption{\small DNN scheduling problem formulation with \sys. \sys takes 1) DNN layer dimensions and 2) DNN accelerator parameters and expresses the scheduling problem into a constrained optimization problem to produce a performant schedule in one shot. }
      \label{fig:overview}
  \end{minipage}
  \hfill
  \begin{minipage}{0.29\textwidth}
      \begin{lstlisting}[language=C++, caption={\small
      An example schedule using the loop
      nest representation for a DNN layer of dimension $R=S=3, P=Q=28, C=8, K=4, N=3$. 
      Same variable prefix indicates tiles from the same problem dimension. },captionpos=b,floatplacement=t,label={lst:example_schedule}]
//DRAM level
for q2 = [0 : 2) :
 // Global Buffer level
 for p2 = [0 : 7) :
  for q1 = [0 : 7) :
   for n0 = [0 : 3) :   
    spatial_for r0 = [0 : 3) :
     spatial_for for k1 = [0 : 2) :
      // Input Buffer level
      spatial_for k0 = [0 : 2) :
       // Weight Buffer level
       for c1 = [0 : 2) :
        for p1 = [0 : 2) :
         // Accumulation Buffer level
         for s0 = [0 : 3) :
          for p0 = [0 : 2) :
           spatial_for c0 = [0 : 8) :
            // Register
            for q0 = [0 : 2) :
      \end{lstlisting}
  \end{minipage}
\vspace{-10pt}
\end{figure*} 
\smallskip

Given that the scheduling space for a DNN layer can have billions of
valid schedules, finding a good schedule through exhaustive search can become an intractable
problem. Table~\ref{table:related_work} shows some recent efforts to
tackle this complexity.

\subsubsection{Brute-force Approaches}
Recent efforts combine exhaustive search
with heuristics to manually prune the scheduling space~\cite{timeloop2019-ispass,
dave2019dmazerunner, interstellar-asplos2020, tillet2019triton,chatarasi2020marvel}. 
To lower the cost of exhaustive search, schedulers in this category typically use a lightweight analytical model to estimate latency, throughput, and power
consumption to compare all valid mappings of a given layer to find the best schedule.
The disadvantages of this approach are two-fold. 
First, such a brute-force
search tends to be exceedingly expensive for complex hardware architectures, making it infeasible to find a good schedule quickly. 
Second, the generated schedules often do not perform optimally since analytical models may fail to consider the communication latency across the spatial hardware. 

\subsubsection{Feedback-based Approaches}
Other recent efforts use feedback-driven approaches along with
machine learning  or
other statistical methods ~\cite{tvm2018-osdi, ragan2013halide,
adams2019learning, jia2019beyond, gamma-iccad2020, hegde2021mind}
to improve the accuracy of the cost model and search for the
solution using black-box or gradient-based search.
Although such approaches can potentially learn the distribution of
the scheduling space,  
they typically require a large amount of training data due to their feedback-driven nature.
As a result, these approaches are mainly applicable to post-silicon hardware where performing a large-scale measurement is possible but are not feasible for hardware under development. 

\subsubsection{Constrained-optimization Approaches}
Constrained-optimization problems, in which objective functions are maximized or minimized subject to given sets of constraints,
have demonstrated the ability to solve many
complex large-scale problems in a reasonable time. Such methods have been widely used in architecture and systems
research for instruction scheduling
~\cite{nowatzki2013general,nowatzki2018hybrid,chin2018architecture}, high-level
synthesis~\cite{cong2006efficient}, memory partitioning~\cite{autotm-asplos2020,
ilp-cases2001}~\cite{cong2011automatic}, algorithm selection~\cite{ilp-multiprocessor,
janus-cgo2019},
and program
synthesis~\cite{phothilimthana2014chlorophyll,sketching-pldi2005,superoptimizers-asplos2006,search-ps-cacm,swizzle-asplos2019}. 

In particular, polyhedral transformation has leveraged constrained-optimization-based
approach for auto-vectorization and loop
tiling~\cite{bondhugula2008practical, grosser2011polly,
kong2013polyhedral,park2013predictive,
baghdadi2015pencil, acharya2018polyhedral}. 
Prior work targets general-purpose CPUs and GPUs that run with
fine-grained instructions and hardware-managed cache, as opposed to the software-managed spatial accelerators that we target.
In addition, existing polyhedral-based approaches~\cite{bondhugula2008practical, baghdadi2015pencil, bagehadi2019tiramisu} lack direct support for tile-size optimization.
Instead, they take the tile size as input and apply a transformation based on the given tile size.
Due to this limitation, the tile size decision cannot be co-optimized with other loop transformations, e.g. loop permutation, in one pass, leading to sub-optimal schedules. 

To address the drawbacks of existing approaches and leverage the regularities from the DNN workloads and the accelerator design for optimization, \sys employs constrained optimization to tackle the DNN scheduling problem in one pass. 
\sys presents a unique domain-specific representation for DNN scheduling that better captures the utilization and communication cost and encodes different loop transformations, i.e., tiling size, loop permutation, and spatial mapping decisions, in one formulation. This unified representation
enables us to solve for all three optimizations in one pass and produce efficient schedules for a complex accelerator system with a
multi-level memory hierarchy.

%% file: 3framework.tex
\section{The \sys{} Framework}

To navigate the large scheduling space of DNN accelerators, we develop \sys, a
constrained-optimization-based DNN scheduler to automatically generate high-performance
schedules for spatially distributed accelerators.
\sys{} not only deterministically solves for a good schedule in one pass without the
need for exhaustive search or iterative sampling, but can also be easily applied
to different network layers and hardware architectures.
This section discusses the \sys framework and how \sys formulates the DNN
scheduling problem with mixed-integer programming (MIP).

\input{3.1overview}
\input{3.2variables.tex}

\input{3.3constraints.tex}

\input{3.4objective.tex}
\input{3.5limitation}

%% file: 3.1overview.tex
\subsection{\sys Overview}

\sys{} optimizes operator-level schedules for mapping
DNN layers onto spatial DNN accelerators.
Specifically, \sys formulates the scheduling problem as a constrained-optimization problem with \textit{variables} representing the schedule, \textit{constraints} representing DNN dimensions and hardware parameters, and \textit{objective} functions representing goals, such as maximizing buffer utilization or achieving better parallelism. 
\figureautorefname~\ref{fig:overview} shows the target problem space of \sys{}. \sys{} takes
the specifications of the DNN layers and the underlying spatial accelerator as
input constraints and generates a valid and high-performance schedule based on the objective functions in one pass.

\subsubsection{Target Workload}
The work targets the DNN operators that can be expressed by a nested loop with 7 variables as
loop bounds: $R, S, P, Q, C, K, N$. $R$ and $S$ refer to the convolution kernel
width and height, $P$ and $Q$ refer to the output width and height, $C$ refers to the input channel size, $K$ refers to the output channel size, and $N$ refers to the batch
size, as illustrated in \figureautorefname~\ref{fig:overview}. The convolution operation computes the dot product of the filter size
$R\times S \times C$ of inputs and weights to generate one point in the output. Matrix multiplications can be expressed in this scheme as well.

\subsubsection{Target Architecture}
\sys targets spatial architectures with an array of processing elements (PEs)
connected via an on-chip network and with multiple levels of memory hierarchy, a
commonly adopted architecture template in today's DNN accelerator
designs~\cite{scaledeep, dadiannao, tangram-asplos19, tetris-asplos17,
interstellar-asplos2020, shao2019-micro, chen2019eyeriss, maeri-asplos2018,
centaur-isca2020, sigma-hpca2020, plasticine-isca2017}.


\begin{figure}[t]
  \centering
  \includegraphics[width=0.9\linewidth]{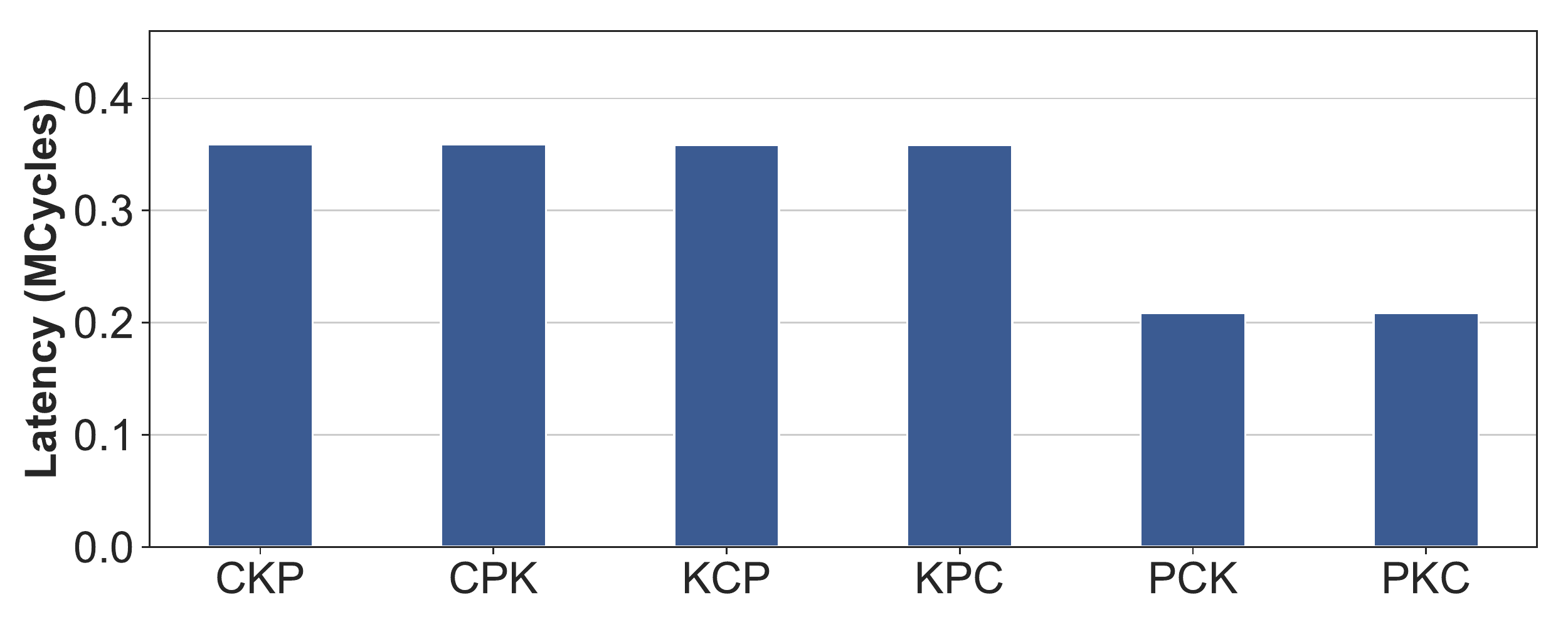}
  \vspace{-7pt}  
  \caption{\small Performance comparison of schedules with different loop permutations for a convolution operator with the layer dimensions of $R=S=3$, $P=Q=8$, $C=32$, $K=1024$. 
The leftmost schedule (\texttt{CKP}) refers to a relative ordering where the input channel dimension (\texttt{C}) is the outermost loop and
the output height dimension (\texttt{P}) is the innermost loop. Since this layer is weight-heavy, loop permutations that emphasize weight reuse, e.g., \texttt{PCK} and \texttt{PKC}, are more efficient.}
  \vspace{-3pt} 
  \label{fig:abls_perm}
\end{figure} 

\begin{figure}[t]
  \centering
  \includegraphics[trim=0 10 0 10, width=\linewidth ]{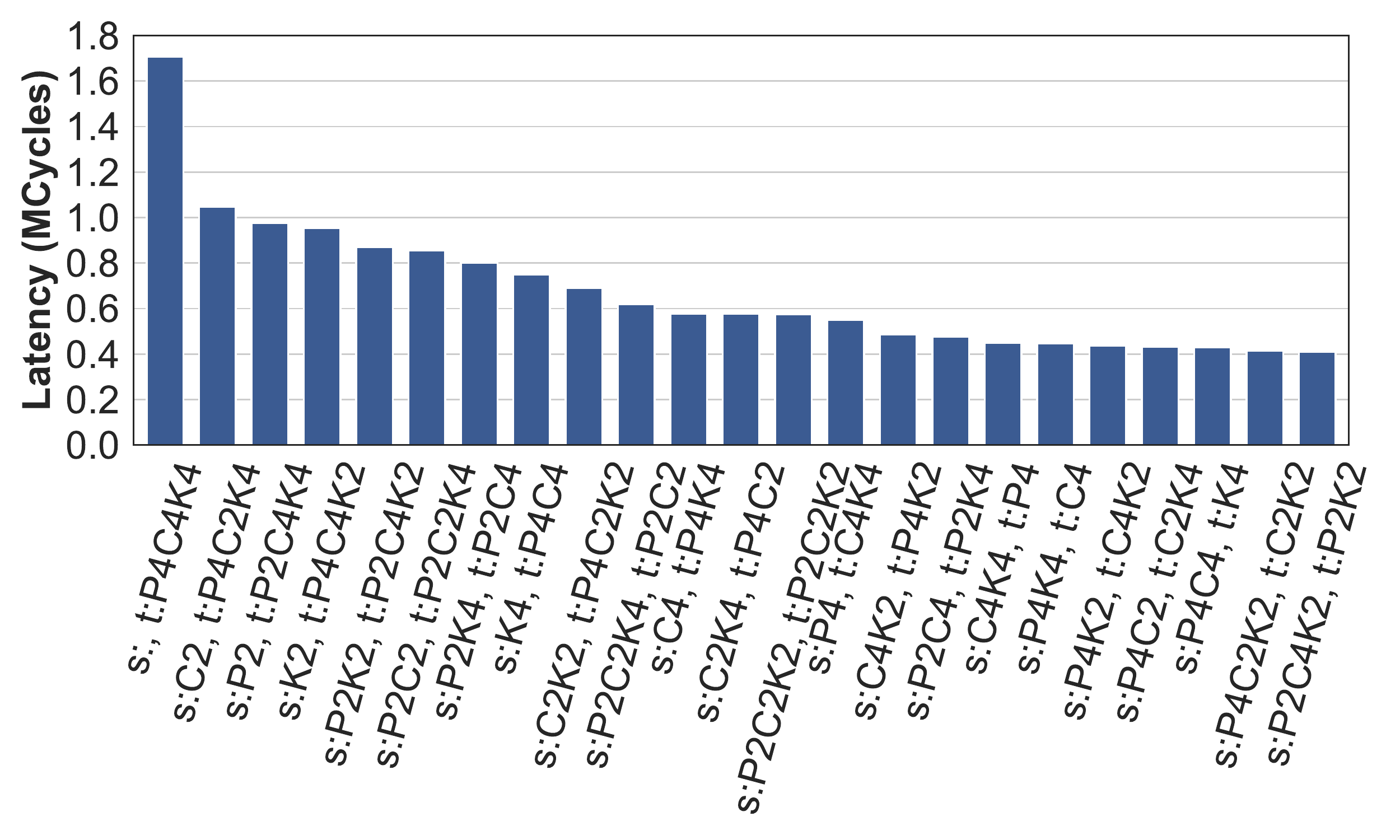}
  \vspace{-17pt}  
  \caption{\small Performance comparison of schedules with different spatial mappings for a convolution operator with the layer dimensions of $R=S=1$, $P=Q=16$, $C=256$, $K=1024$. Factors in \textit{s} list are for spatial mapping, and factors in \textit{t} list are for temporal mapping. For example, \texttt{s:P4C4,t:K4} represents a mapping where a factor 4 of the \texttt{P} dimension and a factor 4 of the \texttt{C} dimension are mapped to spatial execution in a system with 16 PEs, leaving \texttt{K}'s factor 4 to temporal mapping.}
  \label{fig:abls_spatial_perm}
  \vspace{-3pt}   
\end{figure} 

\subsubsection{Target Scheduling Decisions}
\sys-generated schedules describe how a specified DNN
layer is executed on a given spatial architecture.
Listing~\ref{lst:example_schedule} shows an example of a schedule.
Here, we use a loop-nest representation~\cite{timeloop2019-ispass} to explicitly describe how the computation of a convolution layer is
mapped to levels of memory hierarchies.
We highlight three aspects of the schedule:
1) \textbf{loop tiling}, which describes which loops are mapped to which memory level and
the values of the loop bounds;
2) \textbf{loop permutation}, which handles the relative ordering between loops in the same
memory hierarchy; and
3) \textbf{spatial mapping}, which defines which loops are mapped to parallel spatial
resources (shown as \textcolor{blue}{\texttt{spatial\_for}} loops in
Listing~\ref{lst:example_schedule}).
All three factors play a key role in the efficiency of the scheduling
choice.
Next, we highlight the implications of loop permutation and spatial
mapping, both of which are less explored than the well-studied loop tiling.

\figureautorefname~\ref{fig:abls_perm} illustrates the impact of \textbf{loop permutation} for a
convolution layer on a given hardware design.
All the schedules use the same loop tiling and spatial mapping except the loop
ordering at the global-buffer level, as indicated in the labels of the X-axis, where \texttt{CKP} means the input channel dimension (\texttt{C})  is the outermost loop, and
the output height dimension (\texttt{P}) is the innermost loop.
In this case, selecting P as the outermost loop, i.e. \texttt{PCK} and \texttt{PKC},  can lead to a 1.7$\times$
speedup for this layer, motivating the need to consider the implications of loop permutation in the scheduling problem. 


~\figureautorefname~\ref{fig:abls_spatial_perm} shows the impact of \textbf{spatial mapping} on DNN
execution.
We notice that there is a
4.3$\times$ gap between best (rightmost) and worst (leftmost) schedules for the layer in consideration.
The fundamental reason for the differences is the different communication traffic
generated by different spatial mapping options.
The best schedule, i.e., the rightmost schedule in the figure (\texttt{s:P2C4K2, t:P2K2}), is obtained when factors $P=2$, $C=4$, $K=2$ are mapped to the spatial loops, which cannot be achieved by simply choosing either model or data parallelism in the spatial partition.
As a result, a systematic evaluation of different spatial mapping choices is required to find a good schedule.

The rest of the section discusses how \sys formulates the scheduling variables,
constraints, and objectives to solve the DNN scheduling problem.


%% file: 3.2variables.tex
\subsection{\sys{} Variables and Constants}

This section discusses the variables and constants, summarized in Table~\ref{table:notation}, used in \sys{} formulation. 
 \subsubsection{Variable Representation} 
\label{sec:variable_representation}
\begin{table}[h]
\resizebox{1\columnwidth}{!} {
\begin{tabular}{cc|cc|cc}
\toprule
\multicolumn{2}{c|}{ \textbf{\sys Variables} } & \multicolumn{2}{c|}{ \textbf{\sys Constants} } & \multicolumn{2}{c}{ \textbf{Indices} } \\
\midrule
$\mathbf{X}$ & \multirow{4}{*}{ 
\pbox{20cm}{ binary matrix\\to represent\\ a schedule}} & $\mathbf{A}$ & \multirow{2}{*}{\pbox{20cm}{layer dimension to \\ data tensor mapping}} & $i$ & memory level \\
& & & & $j$ & layer dimension \\
& & $\mathbf{B}$ & \multirow{2}{*}{\pbox{20cm}{ memory level to \\ data tensor mapping} } & 
$n$ & prime factor index \\
& & & & $k$ & mapping choice \\
& & & & $z$ & permutation level \\
& & & & $v$ & data tensor\\

\bottomrule
\end{tabular}
}
\vspace{-2pt}
  \caption{\small \sys Notations.}
  \label{table:notation}
\end{table} 

\begin{table}
\small
\begin{tabular}{c} 
\textbf{DNN Layer:} $R=3, S=1, P=1, Q=1, C=1, K=4, N=3$ \\
$\xrightarrow[]{} $ \textbf{Prime Factors:} = [[3],[1],[1],[1],[1],[2,2][3]]
\end{tabular}

\large
\resizebox{\columnwidth}{!} {
\begin{tabular}{c|c|c|c|c|c|c|c|c|c|c|c|c}
\toprule
Idx & \multicolumn{2}{c|}{ } & Perm & \multicolumn{9}{c}{ Schedule }  \\ 
\hline
$j$ & \multicolumn{2}{c|}{ Layer Dim. } & \multirow{3}{*}{ } & \multicolumn{2}{c|}{ R = 3 } & ... & \multicolumn{4}{c|}{ K = 4 } & \multicolumn{2}{c}{ N = 3 } \\ 
\cline{1-3} \cline{4-13}
$n$ & \multicolumn{2}{c|}{ Prime Factors} & & \multicolumn{2}{c|}{ 3 } & ... & \multicolumn{2}{c|}{ 2 } & \multicolumn{2}{c|}{ 2 } & \multicolumn{2}{c}{ 3 } \\
\cline{1-3} \cline{4-13}
$k$ & \multicolumn{2}{c|}{ s / t Mapping } & &s&t&  & s & t & s & t & s & t \\ \midrule
\cline{3-13}
\multirow{8}{*}{ $i$} & \multirow{8}{*}{ \rotatebox[origin=c]{90}{Memory Levels} } & Register & ... &  & & & & & & & &  \\
\cline{3-13}
& ... & ... & & & & & & & & & \\
\cline{3-13}
& & InputBuf & ... & &  & &\cmark& & & & & \\
\cline{3-13}
& & \multirow{5}{*}{ GlobalBuf } & $O_0$ & &  & & & & & & & \\
\cline{4-13}
&& & $O_1$ & & & & & &  & & &\cmark\\
\cline{4-13}
&& & $O_2$ & & &  & & &\cmark& & & \\
\cline{4-13}
&& & ... & & & & & & & & &  \\
\cline{4-13}
&& & $O_{Z}$ &\cmark& & & & & & & &  \\ 
\cline{2-13}
\bottomrule
\end{tabular}
}
\caption{\small Example binary matrix $\mathbf{X}$ representing a schedule. A checkmark in s, t indicates spatial or temporal mapping.
A checkmark in $O_{0},...,O_{Z}$ indicates the rank for loop permutation. In this schedule, the loop tile of size $3$ from problem dimension $N$ is allocated within the GlobalBuf at the innermost loop level, assigned for temporal execution. Both loop tiles from $K$ are  mapped to spatial resources.}
\label{table:schedule}
\end{table}


We devise a mathematical representation for the DNN schedules and formulate the scheduling problem as a prime-factor allocation problem. 
Given a layer specification, we first factorize each loop
bound into its $prime\_factors$.
If the loop bound themselves are large prime number, we can pad them and then factorize. 
We assign each prime factor to a {\em scheduling configuration} that is composed
of a combination of three decisions:
1) the mapped memory level, 2) the permutation order, and 3) the spatial mapping.
Each prime factor has exactly one scheduling configuration.

Here, we use a binary matrix $\mathbf{X}$ to represent the prime factor allocation, i.e., the scheduling space, shown in Table~\ref{table:schedule}.
The four dimensions of $\mathbf{X}$  are: 1) the layer dimension variables (indexed by $j$), 2) the prime factors of the loop bounds (indexed by $n$), 3) whether  it is a spatial or temporal mapping (indexed by $k$), and 4) the memory and the permutation levels (indexed by $i$). 
With the prime factor decomposition, \sys{}'s encoding can represent all
possible schedules and guarantees that the optimization solves for the
full search space. 

Table~\ref{table:schedule} shows an example binary matrix $X$ that represents the schedule shown in Listing~\ref{lst:example_schedule}.
First, \sys{} performs the \textit{tiling} optimizations by assigning the prime factors to different
memory levels.
For example, dimension $K$  is split into two tiles, where the inner tile of size 2 is allocated to the input buffer, and the
outer tile of size 2 is allocated in the global buffer.  
Second, mapping a prime factor to \textit{spatial}
execution is indicated by whether the factor is mapped to
a spatial column $s$ or a temporal column $t$ in the table. In this example, both prime factors for $K$ are spatially mapped.
Finally, for loop \textit{permutation}, we add rank indices $O_0, O_1, ..., O_Z$ to the memory level of interest, where only one prime factor can be mapped to each rank.
The lowest-ranked factor is allocated to the innermost loop, while the highest-ranked factor is allocated to the outermost loop.
In the example shown in Table~\ref{table:schedule}, the problem dimension N is mapped at the $O_1$ level in the global buffer for
temporal mapping, which means the factor $N=3$ will be assigned rank 1 in the global-buffer level.
Without other
factors in the global-buffer level, factor $N=3$ with the smallest rank will become the innermost loop in permutation. For the ranking of permutation, we reserve enough slots for all prime factors at all memory levels. Not all the slots need to be filled since a prime factor can only be allocated to one memory level.


\subsubsection{Constant Parameters}
\label{sec:framework_constants}

\begin{table}[t]
\footnotesize
  \begin{minipage}{.4\linewidth}
    \centering
    \begin{tabular}{c|c|c|c|c}
    \toprule
    \multirow{2}{*}{ } & \multicolumn{3}{c|}{ Related } & Idx \\
    \cline{2-5}
    & W & IA & OA & $v$ \\
    \midrule
    \hline
    R & \cmark & - & & \multirow{7}{*}{ $j$ } \\ 
    \cline{1-4}
    S & \cmark & - & & \\
    \cline{1-4}
    P & & \cmark & \cmark& \\
    \cline{1-4}
    Q & & \cmark & \cmark & \\
    \cline{1-4}
    C & \cmark & \cmark & & \\
    \cline{1-4}
    K & \cmark & & \cmark & \\
    \cline{1-4}
    N & & \cmark & \cmark & \\
    \hline
    \bottomrule
    \end{tabular}
  \end{minipage}
  \qquad
  \begin{minipage}{.45\linewidth}
    \centering
    \begin{tabular}{c|c|c|c|c}
    \toprule
    \multirow{2}{*}{ } & \multicolumn{3}{c}{ Related } & Idx \\
    \cline{2-5}
    & W & IA & OA & $v$ \\
    \midrule 
    \hline
    Register & \cmark & \cmark & \cmark & \multirow{6}{*}{ $i$ } \\
    \cline{1-4}
    AccBuf & & & \cmark & \\
    \cline{1-4}
    WBuf & \cmark & & & \\
    \cline{1-4}
    InputBuf & & \cmark & & \\
    \cline{1-4}
    GlobalBuf & \cmark & \cmark & & \\
    \cline{1-4}
    DRAM & \cmark & \cmark & \cmark & \\
    \hline
    \bottomrule
    \end{tabular}
  \end{minipage}
  \caption{\small Constant binary matrices $\mathbf{A}$ (left) and $\mathbf{B}$
  (right). $\mathbf{A}$ encodes how different layer dimensions associate with
  data tensors. $\mathbf{B}$ encodes which data tensor can be stored in which
  memory hierarchy.}
  \label{tab:constant-matrix}
\end{table}
\smallskip  

In addition to the loop-related variables, we have intrinsic relations
across different components in the architecture
and layer specifications which must be encoded by constant parameters.
\sys uses two constant binary matrices to encode the unique relations
in the DNN scheduling space, shown in Tabel~\ref{tab:constant-matrix}.
The first binary constant matrix, $A$, encodes the association between layer
dimensions (i.e., rows of the matrix) and data tensors (i.e., columns of the matrix).
For each input (IA), weight (W), and output (OA) tensor, matrix $A$ indicates
which layer
dimensions, i.e., $R, S, P, Q, C, K, N$, should be used to calculate the data transaction size as well as
multicast and reduction traffic on the accelerators.

In addition, we introduce another binary matrix $\mathbf{B}$ to
represent which memory hierarchy can be used to store which data tensor.
DNN accelerators typically deploy a multi-level memory hierarchy, where each
memory level can be used to store different types of data tensors.
For example, matrix $\mathbf{B}$ shown in Table~\ref{tab:constant-matrix}
represents an architecture that has dedicated input and weight buffers for
input activation and weight, respectively, while providing a shared global buffer to store input and output activations.

%% file: 3.3constraints.tex
\subsection{\sys{} Constraints} 
\label{sec:constraints}
This section discusses the constraints derived from the target
accelerator architecture that must be satisfied in \sys{} and shows how to
express them with \sys{} variables and constants. 

\subsubsection{Buffer Capacity Constraint}
\label{sec:buf_util}
To generate a valid schedule in a software-managed memory system, a key
constraint is to ensure that the size of data to be
sent to the buffer does not exceed the buffer capacity.
The hardware memory hierarchy can be represented by the binary constant matrix $\mathbf{B}$ discussed earlier.
For each memory buffer, based on the tensor-dimension correlation matrix $\mathbf{A}$,
we calculate the tiling size of each tensor by multiplying the relevant prime factors together indicated by $\mathbf{X}$.
Both spatial and temporal factors
should be included in the buffer utilization. Let $N_j$ be the number of prime factors for the layer dimension $j$. 
Then the utilization of the buffer level $I$ can be expressed
as: 
\begin{equation}
\begin{aligned}
 \prod^{I-1}_{i=0}\prod^{6,\, N_j}_{j=0, n=0}\prod^{1}_{k=0}
\begin{cases}
    prime\_factor_{j,n},& X_{(j,n),i,k} A_{j,v}B_{I,v} = 1\\
    1,              & \text{otherwise}
\end{cases}\\ 
\end{aligned}
\end{equation} 

We then set the upper bound of the buffer utilization to the capacity of different buffer sizes, represented using $M_{I,v}$.
However, a problem with this
utilization constraint is that it involves products of the decision
variables $\mathbf{X}$, making it nonlinear and infeasible to solve with standard
constraint solvers. 
To address this limitation, we take the logarithm of both
sides of the constraints to obtain a linear expression for the utilization and encode the if-else statement as: 
\begin{equation}
\begin{aligned}
U_{I,v} & =\sum^{I-1}_{i=0}\sum^{6,\, N_j}_{j=0,n=0}\sum^{1}_{k=0} \log(prime\_factor_{j,n})A_{j,v}B_{I,v}X_{(j,n),i,k} \\
 & \le \log(M_{I,v}), \forall I
\end{aligned}
\end{equation} 
To encode different precisions for different data tensors, we add the logarithm of the datatype sizes $precision_v$ to $U_{I,v}$. 
\subsubsection{Spatial Resource Constraint}
Another set of \sys{} constraints is from the limited number of spatial
resources. At the chip level, there is a limited number of PEs.  At the PE
level, there is a limited number of multiply-and-accumulate (MAC) units.
In~\sys{}, once a factor is assigned
to spatial mapping in the configuration, it needs to satisfy: 1) each problem factor can only be mapped to either spatial or
temporal execution, 2) factors that map to spatial execution do not exceed the resource limit in the architecture. 
These two constraints can be expressed in the equations below:
\begin{equation}
\begin{aligned}
\sum^{1}_{k=0} X_{(j,n),i,k} == 1, \forall (j,n), i
\end{aligned}
\end{equation} 
\noindent
\begin{equation}
\begin{aligned}
\sum^{6,\,N_j}_{j=0,n=0}\log(prime\_factor_{j,n})X_{(j,n),I,0} \le \log(S_{I}),  \forall I
\end{aligned}
\end{equation} 
\noindent
where $S_{I}$ is the number of available spatial resources at the level $I$.  
\jenny{do we need to a table to illustrate it or no}


%% file: 3.4objective.tex
\subsection{Objective Functions}
\label{sec:framework_obj}
In this section, we describe the objective functions for \sys. Each objective can be either used individually to optimize a single aspect of performance, e.g., utilization, compute, and communication, or combined with others.

\subsubsection{Utilization-Driven Objective}
High on-chip buffer utilization improves data-reuse opportunity. 
As demonstrated in the prior work~\cite{dinh2020communicationoptimal}, communication lower bounds can be achieved when the tiling block size
is optimized for buffer utilization in a system with one-level cache.
In this work, we formulate a utilization objective that aims
to maximize the buffer utilization of all tensors, so the overall
communication is minimized.
We use the same formulation for the buffer utilization as in \ref{sec:buf_util} and maximize the following linear utilization function: 
\begin{equation}
\begin{aligned}
\hat{Util}= \sum^{I-1}_{i=0}\sum^{2}_{v=0}U_{i,v} 
\end{aligned}
\end{equation} 

Here, maximizing the sum of utilization for all buffer levels and all tensors in the
logarithm form is equivalent to maximizing the geometric mean of the buffer
utilization. Users can also attach weights to the different buffer levels or different data tensors if they want to optimize for the utilization of a
specific level of the memory. 

\subsubsection{Compute-Driven Objective}
\label{sec:comput_obj}
The total number of compute cycles is another factor that affects the quality of schedules.
In this formulation, we multiply all the temporal factors
for the estimated compute cycles in each PE. Intuitively, this objective allows
the constraint solver to exploit the parallelism in the system by mapping more
iterations to the spatial resources than to temporal iterations.
The objective can be expressed as a linear function again with logarithm taken:
\begin{equation}
\begin{aligned}
\hat{Comp} = \sum^{I}_{i=0}\sum^{6,\,N_j}_{j=0,n=0}\log(prime\_factor_{j,n})X_{(j,n),i,1}
\end{aligned}
\end{equation} 
\noindent 

\subsubsection{Traffic-Driven Objective}
Communication latency is a key contributing factor to the performance of spatial architecture.
\sys{} also includes a traffic-driven objective to capture the communication cost.  
Specifically, communication traffic can be decomposed into three terms:
1) data size per transfer,
2) spatial factors of multicast and unicast traffic, and
3) temporal iterations.
Multiplying these three factors will get the total amount of traffic in the network.
Next, we discuss how we capture each of these factors using \sys{}'s representation.


First, similar to the buffer utilization expression, data size per transfer can computed using the allocated prime factors in matrix $\mathbf{X}$, together with the dimension-tensor correlation matrix $\mathbf{A}$, as shown in the equation below:
\begin{equation}
\begin{aligned}
D_v = \sum^{I-1}_{i=0}\sum^{6,\,N_j}_{j=0,n=0}\sum^{1}_{k=0}
\log(prime\_factor_{j,n})A_{j,v}X_{(j,n),i,k}
\end{aligned}
\end{equation}

Second, spatial factors would incur different multicast, unicast, and reduction patterns. 
The dimension-tensor correlation matrix $\mathbf{A}$ discussed in Sec~\ref{sec:framework_constants} can be used to indicate different traffic patters.
Specifically, depending on whether the spatial dimension, indicated by the binary matrix $\mathbf{X}$, is related to the specific tensor in consideration, represented by the constant matrix $\mathbf{A}$, different traffic patterns, e.g., multicast vs. unicast or reduction vs. unicast, would occur.

\begin{figure}[t]
  \centering
  \includegraphics[trim=110 10 490 10, width=0.98\linewidth]{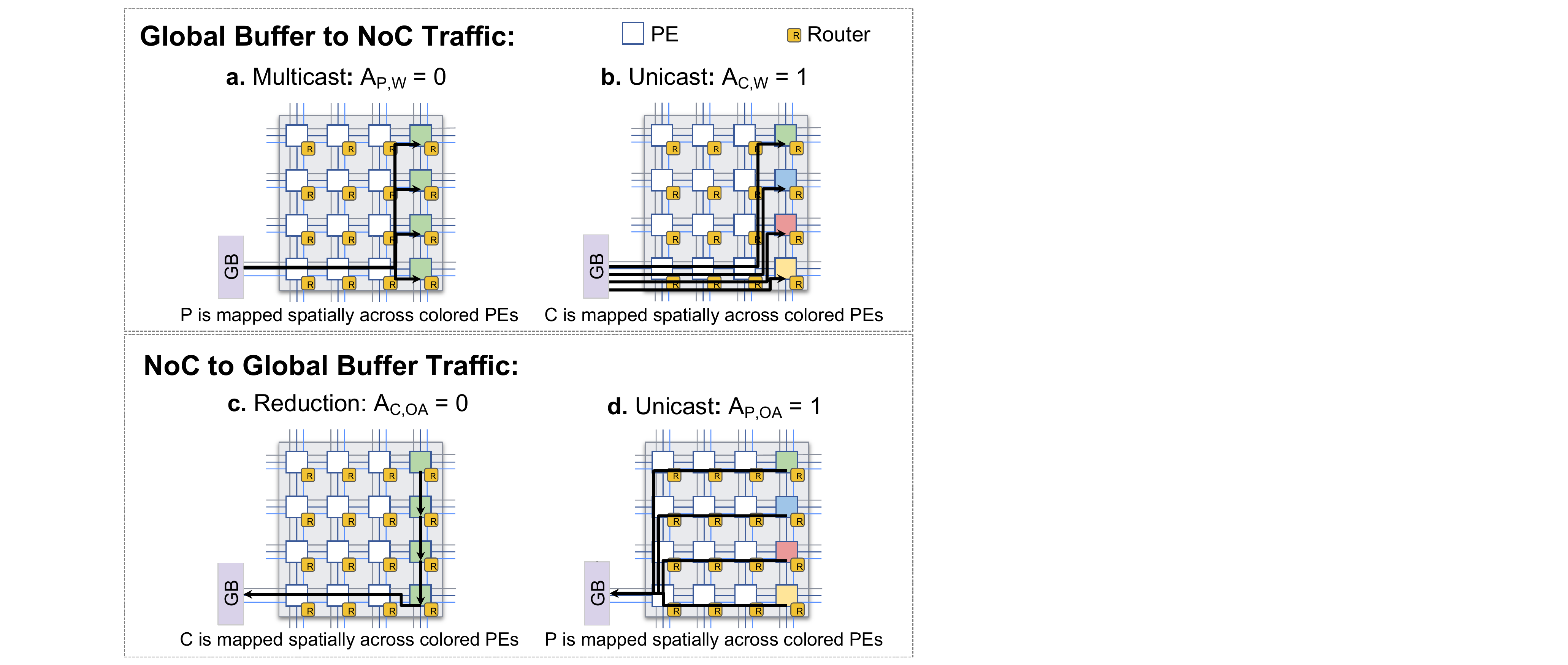}
  \caption{\small 
  Different traffic patterns based on the constant matrix $\mathbf{A}$. The two figures (top) show how the constant $\mathbf{A}$ encodes the traffic types (multicast, unicast, reducation) for different data tensors from the global buffer to PEs. The figures on the bottom show its implication on output tensor reduction traffics. 
  }
  \label{fig:noc_traffic_patterns}
\end{figure} 
 
\figureautorefname~\ref{fig:noc_traffic_patterns} shows how the intrinsic tensor-dimension correlation matrix $\mathbf{A}$ can be used to calculate different traffic patterns for different variables.
For example, as shown in \figureautorefname~\ref{fig:noc_traffic_patterns}a, if the dimension $P$ 
is mapped spatially, $A_{P, \text{W}}=0$ implies multicast traffic for weight tensor W. 
Since weight is not related to $P$, when we send weights from global buffer to PEs, the weight traffic will be multicasted to the destination PEs.
If the dimension $C$ 
is mapped spatially, $A_{C, \text{W}}=1$ (\figureautorefname~\ref{fig:noc_traffic_patterns}b) implies unicast traffic for weight tensor W 
as weight is related to $C$.
Similarly, if the dimension $C$ 
is mapped spatially, 
$A_{C, \text{OA}}=0$  (\figureautorefname~\ref{fig:noc_traffic_patterns}c) implies reduction traffic for output tensor OA, 
where partially sum needs to be reduced across $C$ before sending back to GB.
If the dimension $P$ 
is mapped spatially, 
$A_{P, \text{OA}}=1$  (\figureautorefname~\ref{fig:noc_traffic_patterns}d) would indicate unicast traffic for output tensor OA,  
as each traffic contributes to different regions of the output.
\sys{} formulates this relationship in the following equation:
\begin{equation}
\begin{aligned}
L_v =\sum^{6,\,N_j}_{j=0,n=0}\log(prime\_factor_{j,n})X_{(j,n),I,0}A_{j,v}
\end{aligned}
\end{equation}

The third term, temporal iteration is used to calculate the number of data transfers at the NoC level. 
We introduce a traffic iteration factor $\mathbf{Y}$ that is a function of
$\mathbf{X}$ at the permutation level, $\mathbf{A}$, and $\mathbf{B}$. $\mathbf{Y}$ indicates if the outer NoC loop bound
should be used for different variables. With $\mathbf{Y}$, we ensure that, for
each variable, if a relevant factor term is seen inside the current loop level,
the current loop level's factor should be used to compute the traffic iteration
regardless of whether it is related to the data tensor of the variable of
interest. This is a term that drives the reuse optimization. 
Mathematically, $\mathbf{Y}$ is constrained as: 
\begin{equation}
\begin{aligned}
&Y_{v,z} \ge \sum^{6,\,N_j}_{j=0,n=0}X_{(j,n),z,1}A_{j,v}B_{I,v}, \forall z, \forall v  \\
& Y_{v,z} \ge Y_{v,z-1}, \forall z > 0, \forall v 
\end{aligned}
\end{equation} 
\noindent
\jenny{for simplicity we can ignore these formulations }
Where $z$ represents the position index for permutation and $Z$ equals the total valid levels for permutation. The traffic iteration term can thus be expressed as: 
\begin{equation}
\begin{aligned}
T_v = \sum^{Z-1}_{z=0} \sum^{6,N_j}_{j=0,n=0}\log(prime\_factor_{j,n})Y_{v,z}X_{(j,n),z,1}
\end{aligned}
\end{equation} 
This turns the linear objective into quadratic as we
multiply $\mathbf{Y}$ with $\mathbf{X}$ to indicate whether there is a factor at
the current permutation level.

After we calculate each individual term, we can combine them together for each tensor that contributes to the total traffic in the network. Similar to the logarithmic transformation we did earlier, instead of multiplying these three terms together, we take the logarithm on both sides to get a linear expression of the traffic, as shown in the equation below: 
\begin{equation}
\begin{aligned}
\hat{Traf} = \sum^{2}_{v=0} ( D_v + L_v + T_v )
\end{aligned}
\end{equation} 
\noindent 

\subsubsection{Overall Objective}

One can construct a composite objective comprised of a linear combination of $\hat{Util}$, $\hat{Comp}$, and $\hat{Traf}$, where we want to minimize the compute and communication latency while maximizing the on-chip buffer utilization: 
\begin{equation}
\begin{aligned}
\hat{O}=  - w_U \hat{Util} + w_C\hat{Comp} + w_T \hat{Traf}
\end{aligned}
\label{eqn:obj}
\end{equation} 
where $w_U, w_T, w_C$ are user-selected parameters controlling the importance of each objective. For a system with double-buffering optimization, $w_T$ can be set to map the traffic sizes to the cycles for memory accesses. This brings $w_T \hat{Traf}$ to be of the same importance as  $w_C \hat{Comp}$ in the optimization. 
Another formulation of the overall objective function to balance the memory access and compute cycles is to minimize the difference of the two terms: $\hat{D} = w_T\hat{Traf} - w_C \hat{Comp}$.
The weights of different objectives can be determined by using a set of micro-benchmarks that characterize the compute, memory, and communication latencies of the target architecture.

%% file: 3.5limitation.tex
\subsection{Limitation of \sys}


\sys leverages the regularity from both the problem and the architecture space, where it assumes a dense CNN workload and does not exploit the sparsity of the data. 
It also best targets hardware systems with deterministic behavior and explicitly managed scratchpads. 
This is because, in systems with non-deterministic behaviors, it can be challenging to construct optimization objectives that capture the impact of such behaviors.
However, \sys can be augmented with 
an iterative search on the objective functions and their corresponding hyperparameters to approximate the unknown hardware performance model and directly prune off the invalid points from the search space. 

%% file: 4methodology.tex
\section{Methodology}
\label{sec:methodology} 

\begin{table*}[t]

\centering
\resizebox{0.7\linewidth}{!} {
\begin{tabular}{cc|cc|cc}
\toprule
\multicolumn{2}{l|}{ \textit{Arithmetic :}} & \multicolumn{2}{l|}{ \textit{Storage :} } & \multicolumn{2}{l}{ \textit{Network :} } \\
\cmidrule(lr){1-2}\cmidrule(lr){3-4} \cmidrule(lr){5-6}

\textbf{MACs} & 64 / PE & \textbf{Registers} & 64B / PE & \textbf{Dimension} & 4$\times$4 \\
\textbf{Weight/Input} & \multirow{2}{*}{ 8bit } & \textbf{Accum. Buffer} & 3KB / PE & \textbf{Router} & Wormhole \\
\textbf{Precision} & & \textbf{Weight Buffer} & 32KB / PE & \textbf{Flit Size} & 64b \\
\textbf{Partial-Sum} & \multirow{2}{*}{ 24bit } &\textbf{ Input Buffer} & 8KB / PE & \textbf{Routing} & X-Y \\
\textbf{Precision} & &\textbf{ Global Buffer} & 128KB & \textbf{Multicast} & Yes \\
\bottomrule
\end{tabular}
}
\caption{\small The baseline DNN accelerator architecture.}
\label{table:arch}
\end{table*}
This section discusses the evaluation platforms we use 
followed by the experimental setup for \sys evaluation.

\subsection{Evaluation Platforms} 
\label{sec:infra}
We evaluate the schedules generated by \sys{} on two platforms: 1) Timeloop for cycle performance and energy consumption, and 2) our cycle-exact NoC simulator for overall latency performance.
The latter more accurately captures the communication overhead and concurrent hardware behaviors on a spatial architecture. 

\textbf{Timeloop} provides microarchitecture and technology-specific energy models for estimating the performance and energy on DNN accelerators.
Timeloop reports the performance in terms of the maximum cycles required for each processing element to complete the workload and to perform memory accesses, assuming perfect latency hiding with double buffering. 
The energy consumption in Timeloop is calculated by multiplying the access count on each hardware component with the energy per access and summing the products up. 
The access count is inferred from the schedule and the energy per access is provided by an energy reference table in Timeloop.  

\textbf{NoC Simulator} augments the Timeloop analytical compute model for PEs with a synthesizable NoC implementation to reflect the communication cost. 
Communication is one of the key contributing factors for latency in a NoC-based system, especially for the communication bound schedules.  
The NoC simulator is transaction-based and cycle-exact for modeling the on-chip traffic. 
Leveraging the synthesizable SystemC router design from Matchlib~\cite{khailany2018modular} that supports  unicast and multicast requests, we construct a resizable 2-D mesh network and implement an X-Y routing scheme. 
The simulator captures both computation and communication latencies by concurrently modeling data transfers in the NoC, the PE executions, and off-chip DRAM accesses based on the DRAMSim2 model~\cite{rosenfeld2011dramsim2}, where the impact of traffic congestion on the NoC can also be manifested.


\subsection{Baseline Schedulers}

We evaluate \sys with respect to two other scheduling schemes: 1) a \textbf{Random} scheduler that searches for five different valid schedules, from which we choose the one with the best result for the target metric, and 2) the \textbf{Timeloop Hybrid} mapper in Timeloop~\cite{timeloop2019-ispass} that randomly selects a tiling factorization, prunes superfluous permutations, and then linearly explores the pruned subspace of mappings before it proceeds to the next random factorization. For this mapper, we keep the default termination condition  where each thread self-terminates after visiting 500 consecutive mappings that are valid yet sub-optimal. The mapper is run with 32 threads, each of which independently searches the scheduling space until its termination condition is met. Once all threads have terminated, Timeloop returns the best schedule obtained from all 16,000+ valid schedules.

\subsection{Experiment Setup}
\textbf{Mixed-Integer Program (MIP) Solver:}
\sys uses Gurobi~\cite{gurobi}, a general-purpose optimization solver for MIP and other constrained programming, as the solver.
We specify the \sys variables, constraints, and objective functions
before we invoke the solver. 
The solver takes at most seconds to return a schedule for DNN layers.

\joshil{Details on how we measure time-to-solution?}


\joshil{Specifics of the Networks and Operators: 1D 2D 3D depth-wise, etc}

\begin{figure*}[t]
  \centering
  \includegraphics[width=\linewidth]{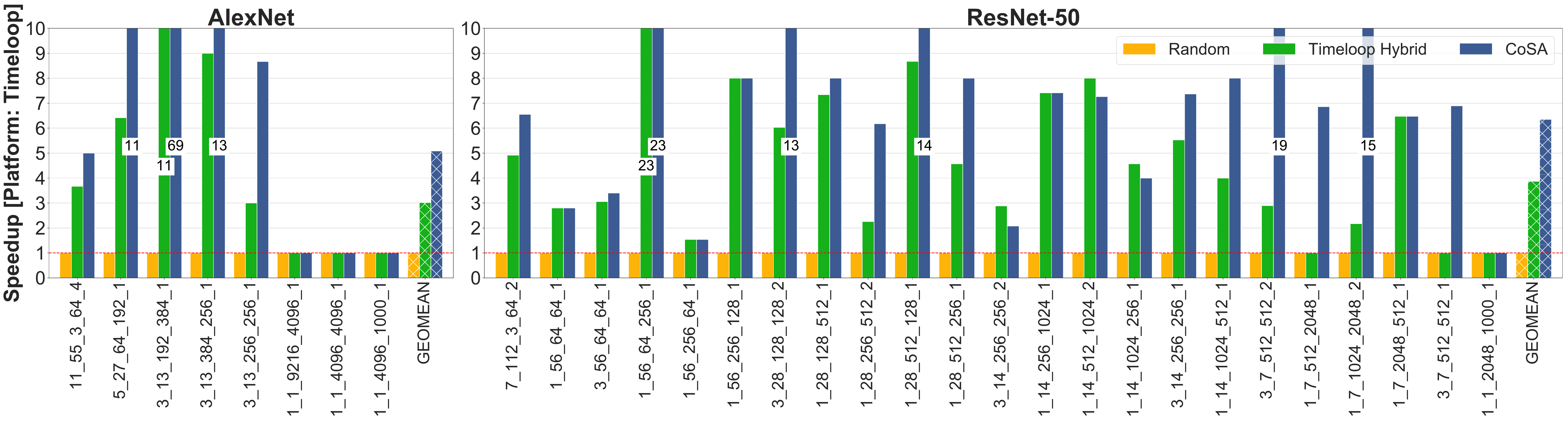}
\includegraphics[width=\linewidth]{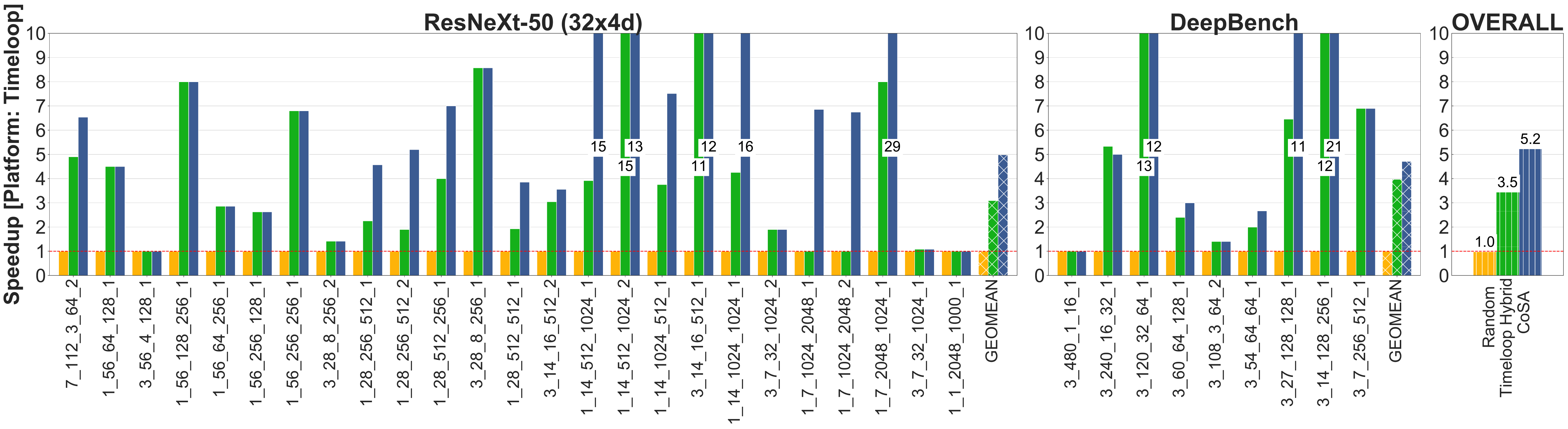}
  \caption{\small Speedup of different schedules relative to Random search on the baseline 4$\times$4 NoC architecture. X-axis labels follow the naming convention \texttt{R\_P\_C\_K\_Stride} where $S=R$ and $Q=P$ in all workloads.  \sys achieves $5.2\times$ and $1.5\times$ higher geomean speedup across four DNN workloads compared to the Random and Timeloop Hybrid search.}
  \label{fig:4x4results_cycle}
\end{figure*} 
\vspace{1ex}

\textbf{DNN workloads:} We measure the performance of \sys-generated schedules over a wide range of DNN workloads targeting different DNN tasks with diverse layer dimensions, including: 
ResNet-50~\cite{resnet}, ResNeXt-50 (32x4d)~\cite{Xie2016resnext},  and Deepbench~\cite{deepbench} (OCR and Face Recognition). 
The precision used for the benchmarks is 8-bit for the input and weights, and 24-bit for the partial sums.
We do not pad the dimensions to be multiples of 2, as it incurs more overhead and outweighs the benefits it provides to allow more scheduling options.


\textbf{Baseline architecture:} We consider a spatial-array architecture like Simba~\cite{shao2019simba} as our baseline. Detailed specifications of the hardware constructs are summarized in Table~\ref{table:arch}.
We demonstrate that the \sys framework is general to be applied for different architecture parameters while delivering high-performance scheduling options in one shot.

%% file: 5evalution.tex
\section{Evaluation}
\label{sec:eval}

In this section, we demonstrate the improved time-to-solution, performance, and energy of \sys compared to baseline schedulers, across different evaluation platforms and different DNN architectures on a diverse set of DNN layers. 

\subsection{Time to Solution}
We compare the average time for \sys and the baseline schedulers to generate the schedule of each layer from the four target DNN workloads.
Table~\ref{tab:time-to-solution} shows that \sys's optimization-driven approach offers more than 90$\times$ (4.2s vs. 379.9s) time-to-solution advantage over the Timeloop Hybrid search strategy.
Timeloop Hybrid search sampled 67 million schedules per layer and evaluated more than 16 thousand valid ones among them, leading to a long runtime.
With Random search, a random sampling of 20K samples in 4.6 seconds resulted in only five valid schedules, further demonstrating the need to have a constraint-based strategy to prune the invalid search space directly. In the following section, we show that \sys not only shortens the time-to-solution but also generates high-quality schedules.

\begin{table}[h]
\centering
\resizebox{\linewidth}{!} {
\begin{tabular}{cccc}
\toprule
& \sys{} & Random (5$\times$) & Timeloop Hybrid \\
\midrule
Avg. Runtime / Layer  & \textbf{4.2s} &   4.6s & 379.9s \\
Avg. Samples / Layer & \textbf{1} & 20K & 67M \\ 
Avg. Evaluations / Layer & \textbf{1} & 5 & 16K+ \\
\bottomrule
\end{tabular}
}
  \caption{\small Time-to-solution Comparison. \sys outputs only one valid schedule per layer. \sys's runtime is $1.1\times$ and $90\times$ shorter than the Random and Timeloop Hybrid search, respectively.}
  \label{tab:time-to-solution}
\end{table}

\subsection{Evaluation on Timeloop Performance and Energy Models}
We compare the performance of the Random search, the Timeloop Hybrid mapper, and the \sys scheduler for four different DNN workloads.
The evaluations are based on our baseline architecture described in Table~\ref{table:arch} and the Timeloop evaluation platform mentioned in Section~\ref{sec:infra}.

\begin{figure}[t]
  \centering
  \includegraphics[width=0.98\linewidth]{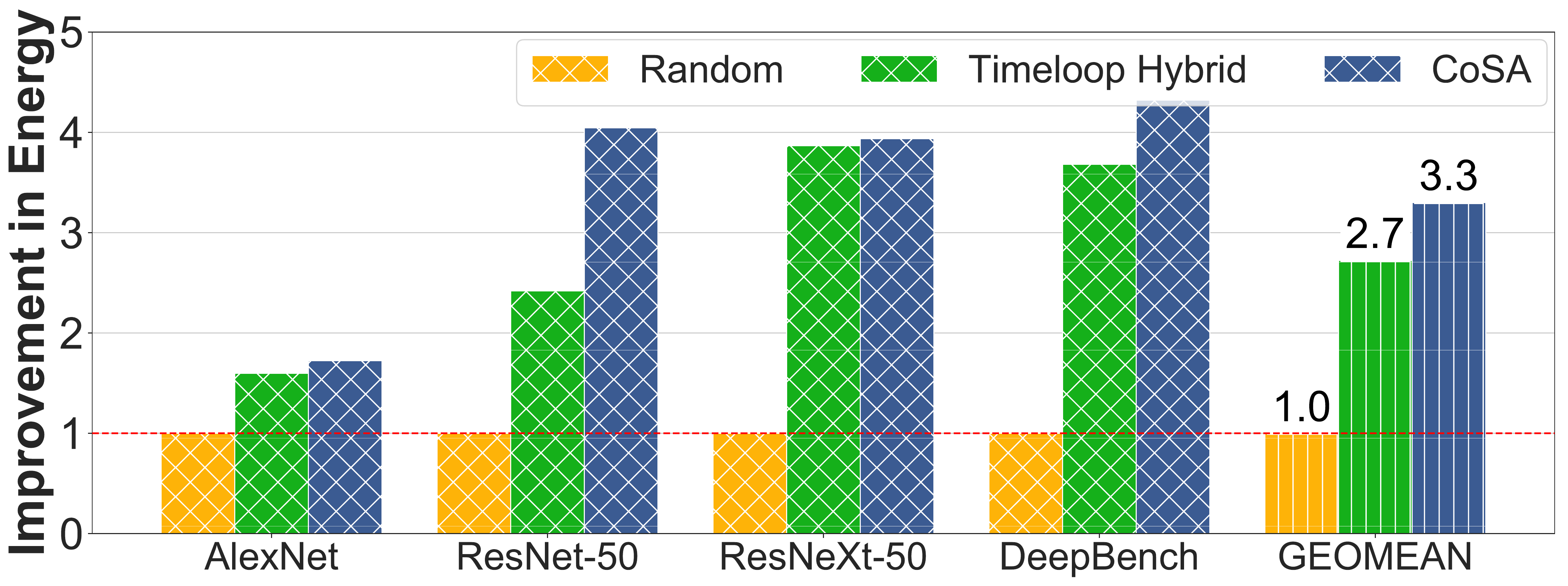}
  \caption{\small Improvements in total network energy reported by the Timeloop energy model. Energy estimations are normalized to results from Random search and are evaluated on the baseline 4$\times$4 NoC.}
  \label{fig:4x4results_energy}
\end{figure} 

\begin{figure}[t]
  \centering
  \includegraphics[width=\linewidth]{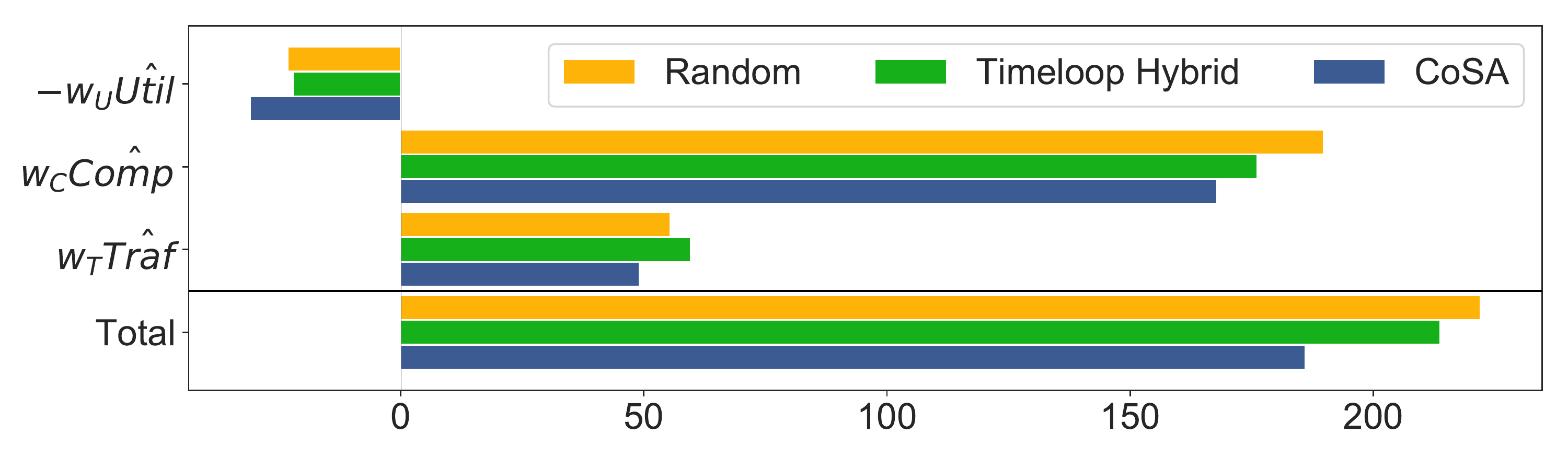}
  \caption{\small  Objective function breakdown for ResNet-50 layer $3\_7\_512\_512\_1$. The goal is to minimize the total objective in Eq.~\ref{eqn:obj}. \sys achieves the lowest values for all objective functions on this layer among all approaches.}
  \label{fig:obj_breakdown}
\end{figure}

\begin{figure*}[t]
  \centering
   \begin{subfigure}[b]{0.98\columnwidth}
    \includegraphics[width=\linewidth]{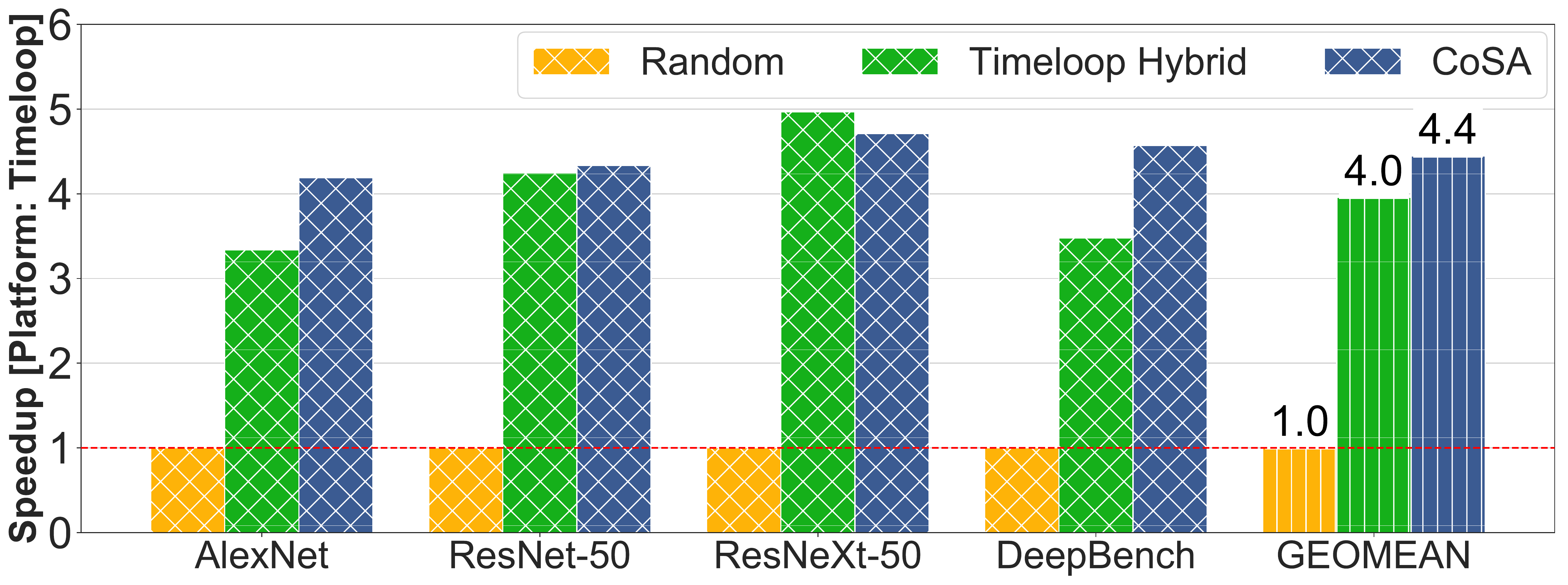}
    \caption{\small $8\times8$ PEs}
    \label{fig:larger_pe}
  \end{subfigure}
   \begin{subfigure}[b]{0.98\columnwidth}
    \includegraphics[width=\linewidth]{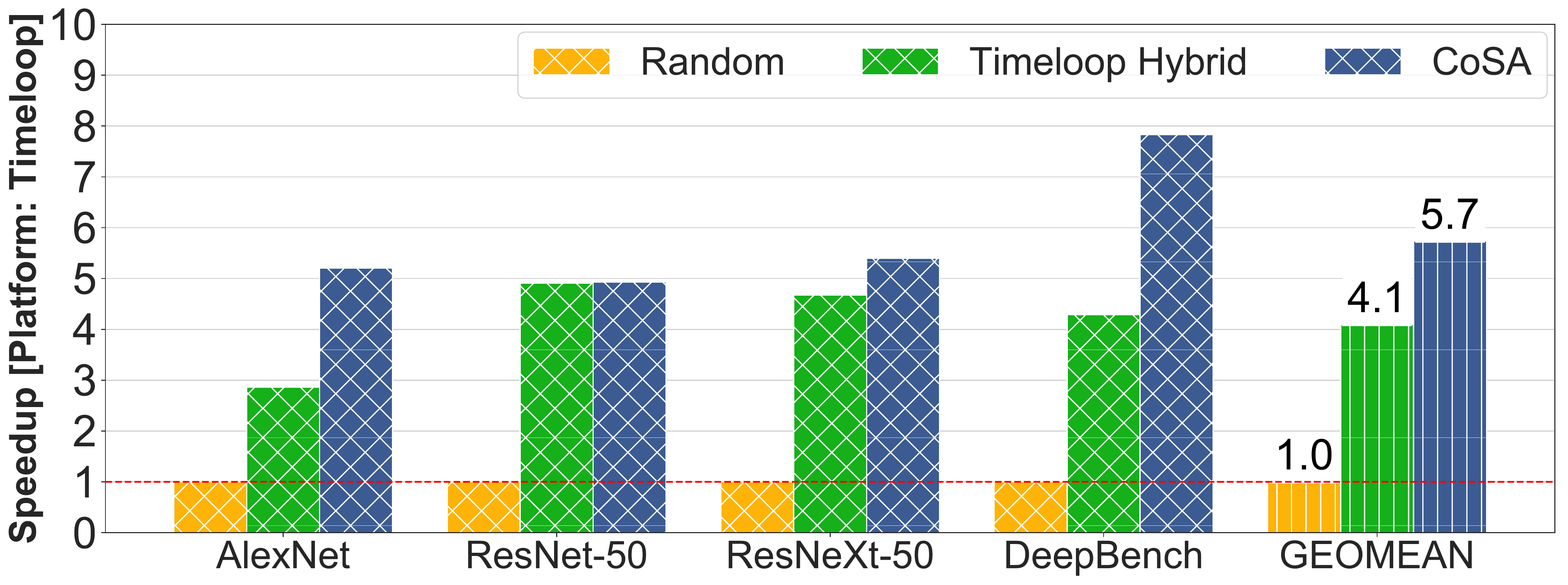}
    \caption{\small Larger Buffers}
    \label{fig:larger_buffer}
  \end{subfigure}
  \caption{\small Speedup relative to Random search reported by Timeloop model on different hardware architectures. \sys's performance generalizes across different hardware architectures with different computing and on-chip storage resources.}
  \label{fig:diffarch}
\end{figure*}
 
 \begin{figure*}[h]
  \centering
  \includegraphics[width=\linewidth]{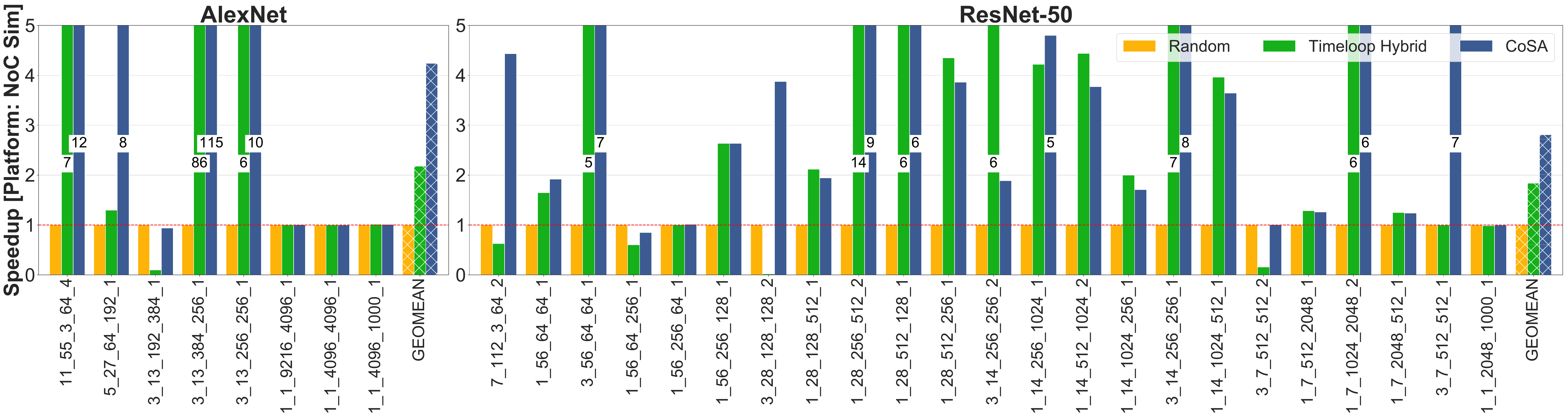}
\includegraphics[width=\linewidth]{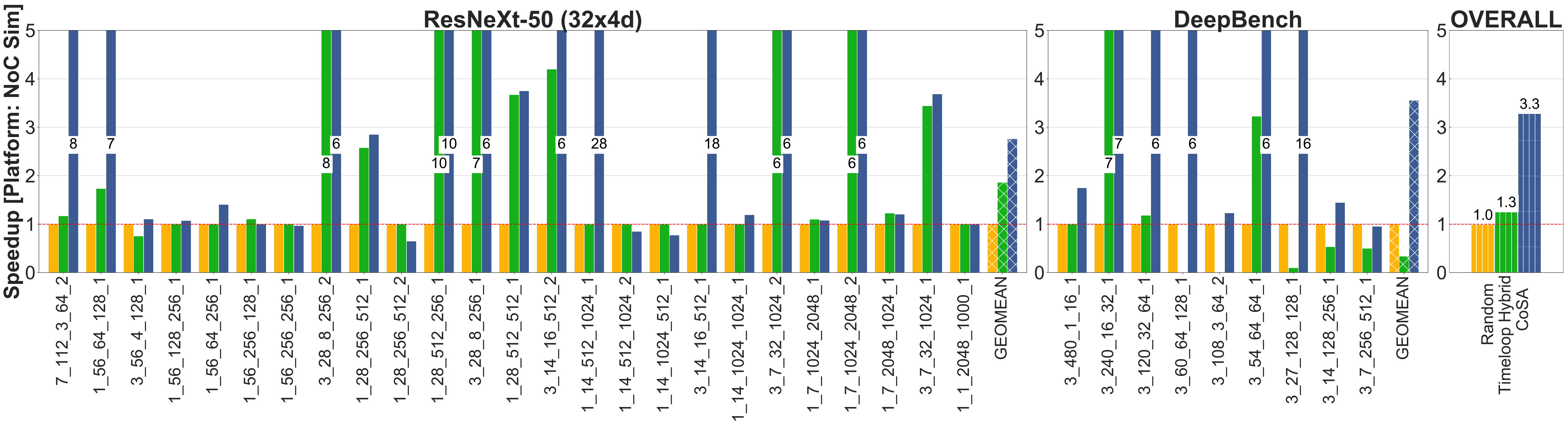}
  \caption{\small Speedup reported by NoC simulator relative to Random search on the baseline 4$\times$4 NoC architecture. 
  \sys achieves $3.3\times$ and $2.5\times$ higher geomean speedup across four DNN workloads compared to the Random and Timeloop Hybrid search on the more communication sensitive NoC simulator.
  }
  \label{fig:4x4results_latency}
\end{figure*} 
\subsubsection{Performance}
\figureautorefname~\ref{fig:4x4results_cycle} shows the speedup reported by Timeloop for different scheduling schemes relative to Random search. 
\figureautorefname~\ref{fig:4x4results_cycle} demonstrates that the \sys-generated schedules are not only valid but also outperform the ones generated by both Random search and Timeloop Hybrid search. 
The geometric mean of the speedups of \sys schedules relative to the Random and Timeloop Hybrid search ones are $5.2\times$ and $1.5\times$ respectively across four DNNs. 

In the few layers where Timeloop Hybrid search slightly outperforms \sys, we find a higher iteration count at the DRAM level in Timeloop Hybrid schedules, which helps to reduce the size of each DRAM transaction and balance the pipeline.
Fine tuning the weights of the objective functions could be used to further improve the \sys-generated schedules.

A more exhaustive Timeloop Hybrid search (32K valid schedules) results in an improvement of only 7.5\%  in latency while increasing runtime by $2\times$. We find that even with $2\times$ more valid samples evaluated, Timeloop Hybrid search still cannot generate schedules that are of similar efficiency to \sys.



\subsubsection{Energy}
We use the Timeloop energy model to evaluate the energy of different schedules. 
Because energy cost is highly correlated with the access count on each hardware component, our traffic objective in \sys is used for the schedule optimization targeting energy efficiency. 
\figureautorefname~\ref{fig:4x4results_energy} demonstrates that \sys, using no simulation feedback, can generate schedules %
22\% more energy-efficient than the best Timeloop Hybrid solutions selected from 16,000+ valid schedules optimizing the energy. 

\subsubsection{Objective Breakdown}
A detailed breakdown of the \sys objective function on ResNet50 layer $3\_7\_512\_512\_1$ is included in \figureautorefname{}\ref{fig:obj_breakdown}. 
Our overall objective function aims to capture an optimization heuristic to maximize the utilization and minimize the compute and traffic costs at the same time with a weighted sum of the three. 
\figureautorefname{}\ref{fig:obj_breakdown} shows that CoSA achieves the lowest total objective among all approaches, and optimizes all three sub-objectives simultaneously.
This observation on the objective values aligns with our empirical results in~\figureautorefname{}~\ref{fig:4x4results_cycle}, where \sys schedule runs $7\times$ faster than the ones generated by Random and Timeloop Hybrid search.

\subsubsection{Different HW Architectures}

We further explore the performance of \sys with different DNN architecture parameters such as different PE array sizes and different SRAM buffer sizes. 
We apply the same weights for the evaluation on the same architecture and customize the objective weights in Eqn.\ref{eqn:obj} using a micro-benchmark for different architectures.
\figureautorefname{}\ref{fig:diffarch} shows the geomean speedup of CoSA across all networks on two different hardware architectures.


\textbf{PE Array Dimension}.
We scale the number of PEs up by $4\times$ and increase both the on-chip communication and DRAM bandwidth by $2\times$ correspondingly. Both of these modifications significantly impact the compute and communication patterns of DNN layer executions. With a larger spatial array of arithmetic units, this case study presents a scheduling problem where decisions about spatial and temporal mapping can be especially crucial to attaining high performance. ~\figureautorefname{}~\ref{fig:larger_pe} shows that CoSA achieves $4.4\times$ and $1.1\times$  speedup compared to Random and Timeloop Hybrid search respectively across four networks. 
This shows that the performance of our scheduler can scale and generalize to NoCs with more PEs, which tend to be more affected by communication costs.  

\textbf{SRAM Size}.
We also increase the sizes of the local and global buffers to demonstrate that \sys can achieve consistently good schedules across different architectures. The sizes of local buffers, i.e. accumulation, weight, and input buffers, are doubled and the global buffer size increased $8\times$. Modified memory capacities, at the PE and global buffer level, are likely to impact the optimal strategy for data re-use and NoC communication traffic reduction. With CoSA, we show  $5.7\times$ speedup over Random and $1.4\times$ speedup over Timeloop Hybrid search in~\figureautorefname{}~\ref{fig:larger_buffer}, demonstrating \sys's capability across different architectures.


\subsection{Evaluation on NoC Simulator}

To further compare the quality of schedules generated by different scheduling schemes, we evaluate them on our NoC simulation platform. 
The NoC simulation platform more accurately captures the communication overhead from the on-chip network as compared to the Timeloop models. 

\figureautorefname~\ref{fig:4x4results_latency} shows the speedup relative to the Random baseline. 
We observe that \sys-generated schedules outperform the baseline schedules for all four DNN workloads, with the greatest performance gains occurring for convolutional layers, e.g. DeepBench layers. Intriguingly, for these same layers, Timeloop Hybrid scheduler actually under-performs Random search as its internal analytical model does not accurately capture the communication traffic in the network.
On the other hand, there is no significant difference between the performance of FC layers among different schedules, as the FC layers are heavily memory-bound with low PE utilization. The DRAM access time dominates in these layers even with the best schedules with respect to reuse of buffered data. 

Overall, \sys achieves a geometric average of up to $3.3\times$ speedup relative to the best Random search solutions and $2.5\times$ relative to Timeloop Hybrid search schedules across the four networks. 
Furthermore, unlike the iterative nature of Random and Timeloop Hybrid search schedules, \sys schedules are consistently performant with the one-shot solution. 





\subsection{Evaluation on GPU}

\begin{figure}[t]
  \centering
  \includegraphics[width=0.98\linewidth]{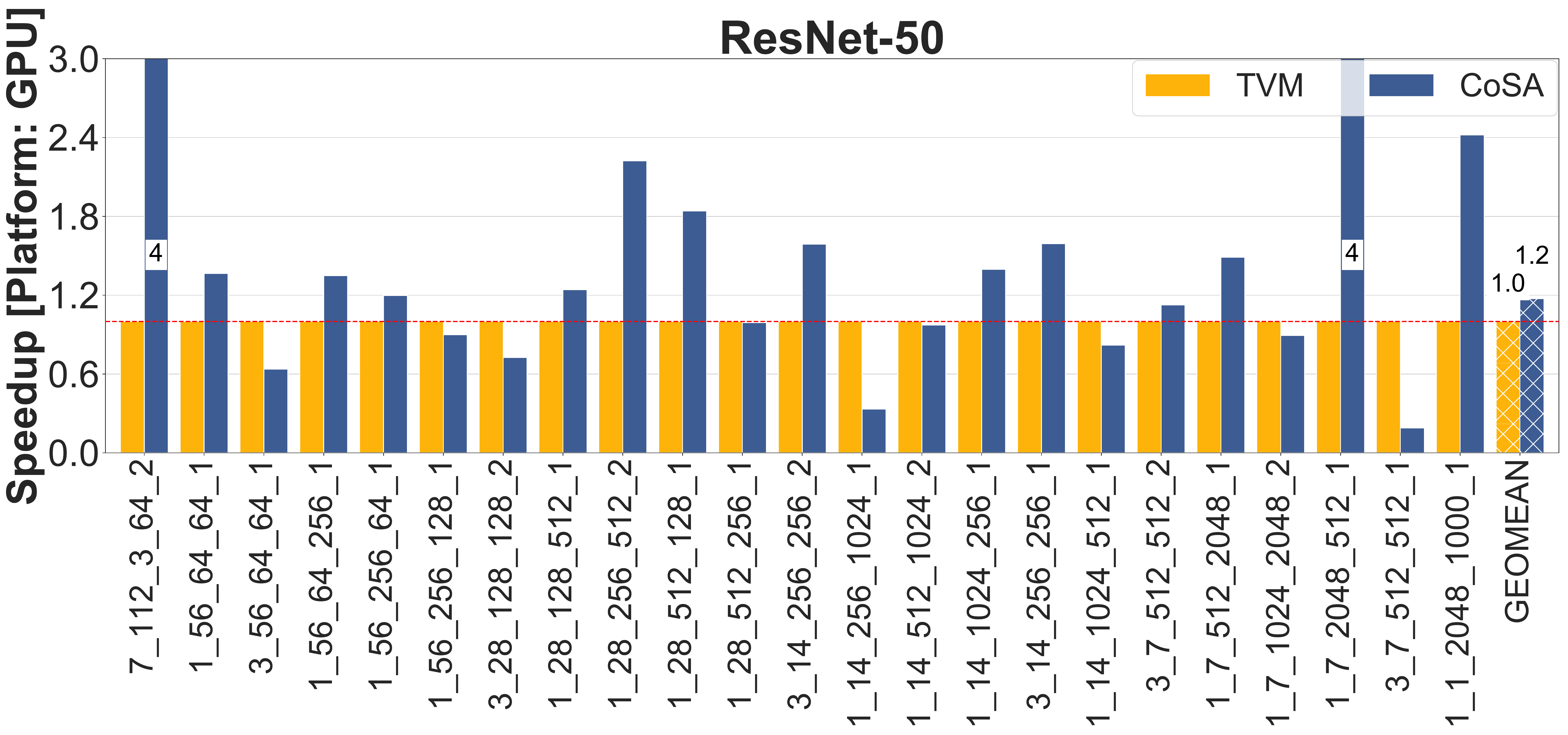}
  \caption{\small Speedup relative to TVM reported on K80 GPU. }
  \label{fig:tvm}
\end{figure}

To show the potential use of \sys for general-purpose hardware, we also 
formulate GPU scheduling as a constrained-optimization problem using \sys. 
We evaluate the performance of \sys on GPU and compare it against TVM~\cite{tvm2018-osdi}.

\textbf{Target GPU.} We target NVIDIA K80 GPU with 2496 CUDA cores and a 1.5MB L2 cache. This GPU has a 48KB shared memory and 64KB local registers, shared by a maximum of 1024 threads in each CUDA thread block.
The thread block is a programming abstraction that represents a group of threads that can be run  serially or in parallel in CUDA. 
The maximum dimension of a thread block is (1024, 1024, 64). Violating these constraints in the CUDA kernel results in invalid schedules. 

\textbf{Constraints.} \sys expresses the hardware constraints for GPU thread groups and shared/local memory similarly to how we specify the spatial resource and buffer capacity constraints in Section~\ref{sec:constraints}. Each thread group can be seen as a spatial level with a specific size. The product of all three thread group sizes is enforced to be smaller than 1024. The share memory utilization is calculated as buffer capacity constraints, and the register utilization is calculated by multiplying the total number of threads with the inner loop register utilization.    

\textbf{Objective Functions.}  
In \sys, we compute the compute objective by discounting the total compute cycles with the total number of threads for GPU, to reflect the performance gain from thread-level parallelism. 
We then adjust the weights of the other objectives using a micro-benchmark. 

We run TVM with the XGBoost tuner for 50 trials per layer as the baseline. \sys generates valid schedules in one shot with a time-to-solution $2,500\times$ shorter than TVM (0.02s vs. 50s per layer). The \sys-generated schedules achieve $1.10\times$ geomean speedup compared to the TVM schedules on ResNet50 as shown in~\figureautorefname{}\ref{fig:tvm}. 

%% file: 6conclusion.tex
\section{Conclusion}
\label{sec:conclusion}

In this paper, we present~\sys{}, an optimization-driven approach to DNN scheduling. 
Harnessing the regularities from DNN workloads and target accelerator designs, we formulate scheduling into a constrained optimization problem that can be solved directly without incurring the high cost of iterative scheduling.
We devise a single mathematical formulation to simultaneously solve for all three key optimizations in scheduling: loop tiling, loop permutation, and spatial mapping. 
Comparing our results to schedules generated from the state-of-the-art work, our approach achieves up to $2.5\times$ speedup and $22\%$ better energy-efficiency,  with $90\times$ shorter time-to-solution. 



%% file: 8ack.tex
\section*{Acknowledgements}
The authors would like to thank Lianmin Zheng for providing the TVM tuning scripts and scheduling templates,
and Kostadin Ilov for the computing system support. 
This work was supported in part 
by the CONIX Research Center, one of six centers in JUMP, a Semiconductor Research Corporation (SRC) program sponsored by DARPA, Berkeley Wireless Research Center, ADEPT Lab industrial sponsors (Intel, Apple, Futurewei, Google, Qualcomm, Seagate, Western Digital), and a Facebook Faculty Research Award.

%% file: main.bbl
\begin{thebibliography}{10}
\providecommand{\url}[1]{#1}
\csname url@samestyle\endcsname
\providecommand{\newblock}{\relax}
\providecommand{\bibinfo}[2]{#2}
\providecommand{\BIBentrySTDinterwordspacing}{\spaceskip=0pt\relax}
\providecommand{\BIBentryALTinterwordstretchfactor}{4}
\providecommand{\BIBentryALTinterwordspacing}{\spaceskip=\fontdimen2\font plus
\BIBentryALTinterwordstretchfactor\fontdimen3\font minus
  \fontdimen4\font\relax}
\providecommand{\BIBforeignlanguage}[2]{{%
\expandafter\ifx\csname l@#1\endcsname\relax
\typeout{** WARNING: IEEEtranS.bst: No hyphenation pattern has been}%
\typeout{** loaded for the language `#1'. Using the pattern for}%
\typeout{** the default language instead.}%
\else
\language=\csname l@#1\endcsname
\fi
#2}}
\providecommand{\BIBdecl}{\relax}
\BIBdecl

\bibitem{tpu_edge}
``{Edge TPU},'' \url{https://cloud.google.com/edge-tpu/}, accessed: 2018-12-05.

\bibitem{acharya2018polyhedral}
A.~Acharya, U.~Bondhugula, and A.~Cohen, ``Polyhedral auto-transformation with
  no integer linear programming,'' in \emph{Proceedings of the ACM SIGPLAN
  Conference on Programming Language Design and Implementation (PLDI)}, 2018.

\bibitem{adams2019learning}
A.~Adams, K.~Ma, L.~Anderson, R.~Baghdadi, T.-M. Li, M.~Gharbi, B.~Steiner,
  S.~Johnson, K.~Fatahalian, F.~Durand, and J.~Ragan-Kelley, ``Learning to
  optimize halide with tree search and random programs,'' \emph{ACM
  Transactions on Graphics (TOG)}, 2019.

\bibitem{search-ps-cacm}
R.~Alur, R.~Singh, D.~Fisman, and A.~Solar-Lezama, ``Search-based program
  synthesis,'' \emph{Communications of the ACM}, 2018.

\bibitem{aws-inferentia}
Amazon, ``{AWS Inferentia: High Performance Machine Learning Inference Chip},''
  \url{https://aws.amazon.com/machine-learning/inferentia/}, 2018.

\bibitem{ansel2014opentuner}
J.~Ansel, S.~Kamil, K.~Veeramachaneni, J.~Ragan-Kelley, J.~Bosboom, U.-M.
  O'Reilly, and S.~Amarasinghe, ``Opentuner: An extensible framework for
  program autotuning,'' in \emph{Proceedings of the International Conference on
  Parallel Architectures and Compilation Techniques (PACT)}, 2014.

\bibitem{ilp-cases2001}
O.~Avissar, R.~Barua, and D.~Stewart, ``Heterogeneous memory management for
  embedded systems,'' in \emph{Proceedings of the International Conference on
  Compilers, Architecture, and Synthesis for Embedded Systems}, 2001.

\bibitem{bagehadi2019tiramisu}
R.~{Baghdadi}, J.~{Ray}, M.~B. {Romdhane}, E.~D. {Sozzo}, A.~{Akkas},
  Y.~{Zhang}, P.~{Suriana}, S.~{Kamil}, and S.~{Amarasinghe}, ``Tiramisu: A
  polyhedral compiler for expressing fast and portable code,'' in
  \emph{International Symposium on Code Generation and Optimization (CGO)},
  2019.

\bibitem{baghdadi2015pencil}
R.~Baghdadi, U.~Beaugnon, A.~Cohen, T.~Grosser, M.~Kruse, C.~Reddy,
  S.~Verdoolaege, A.~Betts, A.~F. Donaldson, J.~Ketema, J.~Absar,
  S.~Van~Haastregt, A.~Kravets, A.~Lokhmotov, R.~David, and E.~Hajiyev,
  ``Pencil: A platform-neutral compute intermediate language for accelerator
  programming,'' in \emph{Proceedings of the International Conference on
  Parallel Architectures and Compilation Techniques (PACT)}, 2015.

\bibitem{superoptimizers-asplos2006}
S.~Bansal and A.~Aiken, ``Automatic generation of peephole superoptimizers,''
  in \emph{Proceedings of the International Conference on Architectural Support
  for Programming Languages and Operation Systems (ASPLOS)}, 2006.

\bibitem{drivenet}
M.~Bojarski, P.~Yeres, A.~Choromanska, K.~Choromanski, B.~Firner, L.~Jackel,
  and U.~Muller, ``Explaining how a deep neural network trained with end-to-end
  learning steers a car,'' 2017.

\bibitem{bondhugula2016pluto+}
U.~Bondhugula, A.~Acharya, and A.~Cohen, ``The pluto+ algorithm: A practical
  approach for parallelization and locality optimization of affine loop
  nests,'' \emph{ACM Transactions on Programming Languages and Systems
  (TOPLAS)}, 2016.

\bibitem{bondhugula2008practical}
U.~Bondhugula, A.~Hartono, J.~Ramanujam, and P.~Sadayappan, ``A practical
  automatic polyhedral parallelizer and locality optimizer,'' in
  \emph{Proceedings of the ACM SIGPLAN Conference on Programming Language
  Design and Implementation (PLDI)}, 2008.

\bibitem{chatarasi2020marvel}
P.~Chatarasi, H.~Kwon, N.~Raina, S.~Malik, V.~Haridas, A.~Parashar,
  M.~Pellauer, T.~Krishna, and V.~Sarkar, ``Marvel: A data-centric compiler for
  dnn operators on spatial accelerators,'' 2020.

\bibitem{tvm2018-osdi}
T.~Chen, T.~Moreau, Z.~Jiang, L.~Zheng, E.~Yan, H.~Shen, M.~Cowan, L.~Wang,
  Y.~Hu, L.~Ceze, C.~Guestrin, and A.~Krishnamurthy, ``{TVM: An Automated
  End-to-end Optimizing Compiler for Deep Learning},'' in \emph{USENIX
  Symposium on Operating Systems Design and Implementation (OSDI)}, 2018.

\bibitem{diannao}
T.~Chen, Z.~Du, N.~Sun, J.~Wang, C.~Wu, Y.~Chen, and O.~Temam, ``{DianNao: A
  Small-footprint High-throughput Accelerator for Ubiquitous
  Machine-learning},'' in \emph{Proceedings of the International Conference on
  Architectural Support for Programming Languages and Operation Systems
  (ASPLOS)}, March 2014.

\bibitem{eyeriss-isca2016}
Y.-H. Chen, J.~Emer, and V.~Sze, ``{Eyeriss: A Spatial Architecture for
  Energy-efficient Dataflow for Convolutional Neural Networks},'' in
  \emph{Proceedings of the International Symposium on Computer Architecture
  (ISCA)}, 2016.

\bibitem{chen2019eyeriss}
Y.-H. Chen, T.-J. Yang, J.~Emer, and V.~Sze, ``Eyeriss v2: A flexible
  accelerator for emerging deep neural networks on mobile devices,'' \emph{IEEE
  Journal on Emerging and Selected Topics in Circuits and Systems}, 2019.

\bibitem{dadiannao}
Y.~Chen, T.~Luo, S.~Liu, S.~Zhang, L.~He, J.~Wang, L.~Li, T.~Chen, Z.~Xu,
  N.~Sun, and O.~Temam, ``{DaDianNao: A Machine-learning Supercomputer},'' in
  \emph{Proceedings of the International Symposium on Microarchitecture
  (MICRO)}, 2014.

\bibitem{chin2018architecture}
S.~A. Chin and J.~H. Anderson, ``An architecture-agnostic integer linear
  programming approach to cgra mapping,'' in \emph{Design Automation Conference
  (DAC)}, 2018.

\bibitem{cong2011automatic}
J.~Cong, W.~Jiang, B.~Liu, and Y.~Zou, ``Automatic memory partitioning and
  scheduling for throughput and power optimization,'' \emph{ACM Transactions on
  Design Automation of Electronic Systems (TODAES)}, 2011.

\bibitem{cong2006efficient}
J.~Cong and Z.~Zhang, ``An efficient and versatile scheduling algorithm based
  on sdc formulation,'' in \emph{Design Automation Conference (DAC)}, 2006.

\bibitem{dave2019dmazerunner}
S.~Dave, Y.~Kim, S.~Avancha, K.~Lee, and A.~Shrivastava, ``{DMazeRunner}:
  Executing perfectly nested loops on dataflow accelerators,'' \emph{ACM
  Transactions on Embedded Computing Systems}, 2019.

\bibitem{deepbench}
DeepBench, ``http://www.github.com/baidu-research/deepbench.''

\bibitem{dinh2020communicationoptimal}
G.~Dinh and J.~Demmel, ``Communication-optimal tilings for projective nested
  loops with arbitrary bounds,'' 2020.

\bibitem{shidiannao-isca2015}
Z.~Du, R.~Fasthuber, T.~Chen, P.~Ienne, L.~Li, T.~Luo, X.~Feng, Y.~Chen, and
  O.~Temam, ``{ShiDianNao: Shifting Vision Processing Closer to the Sensor},''
  in \emph{Proceedings of the International Symposium on Computer Architecture
  (ISCA)}, 2015.

\bibitem{brainwave-isca-2018}
J.~Fowers, K.~Ovtcharov, M.~Papamichael, T.~Massengill, M.~Liu, D.~Lo,
  S.~Alkalay, M.~Haselman, L.~Adams, M.~Ghandi, S.~Heil, P.~Patel, A.~Sapek,
  G.~Weisz, L.~Woods, S.~Lanka, S.~Reinhardt, A.~Caulfield, E.~Chung, and
  D.~Burger, ``{A Configurable Cloud-Scale DNN Processor for Real-Time AI},''
  in \emph{Proceedings of the International Symposium on Computer Architecture
  (ISCA)}, 2018.

\bibitem{tetris-asplos17}
M.~Gao, J.~Pu, X.~Yang, M.~Horowitz, and C.~Kozyrakis, ``{Tetris: Scalable and
  Efficient Neural Network Acceleration with 3D Memory},'' in \emph{Proceedings
  of the International Conference on Architectural Support for Programming
  Languages and Operation Systems (ASPLOS)}, 2017.

\bibitem{tangram-asplos19}
M.~Gao, X.~Yang, J.~Pu, M.~Horowitz, and C.~Kozyrakis, ``{Tangram: Optimized
  Coarse-Grained Dataflow for Scalable NN Accelerators},'' in \emph{Proceedings
  of the International Conference on Architectural Support for Programming
  Languages and Operation Systems (ASPLOS)}, 2019.

\bibitem{grosser2011polly}
T.~Grosser, H.~Zheng, R.~Aloor, A.~Simb{\"u}rger, A.~Gr{\"o}{\ss}linger, and
  L.-N. Pouchet, ``Polly-polyhedral optimization in llvm,'' in
  \emph{Proceedings of the First International Workshop on Polyhedral
  Compilation Techniques (IMPACT)}, 2011.

\bibitem{gupta-dlrm-hpca2020}
U.~{Gupta}, C.~{Wu}, X.~{Wang}, M.~{Naumov}, B.~{Reagen}, D.~{Brooks},
  B.~{Cottel}, K.~{Hazelwood}, M.~{Hempstead}, B.~{Jia}, H.~S. {Lee},
  A.~{Malevich}, D.~{Mudigere}, M.~{Smelyanskiy}, L.~{Xiong}, and X.~{Zhang},
  ``The architectural implications of facebook's dnn-based personalized
  recommendation,'' in \emph{Proceedings of the International Symposium on
  High-Performance Computer Architecture (HPCA)}, 2020.

\bibitem{gurobi}
\BIBentryALTinterwordspacing
L.~Gurobi~Optimization, ``Gurobi optimizer reference manual,'' 2020. [Online].
  Available: \url{http://www.gurobi.com}
\BIBentrySTDinterwordspacing

\bibitem{ilp-multiprocessor}
M.~W. {Hall}, J.~M. {Anderson}, S.~P. {Amarasinghe}, B.~R. {Murphy}, {Shih-Wei
  Liao}, E.~{Bugnion}, and M.~S. {Lam}, ``Maximizing multiprocessor performance
  with the suif compiler,'' \emph{IEEE Computer}, 1996.

\bibitem{resnet}
K.~He, X.~Zhang, S.~Ren, and J.~Sun, ``{Deep Residual Learning for Image
  Recognition},'' in \emph{Proceedings of the Conference on Computer Vision and
  Pattern Recognition (CVPR)}, 2016.

\bibitem{hegde2021mind}
K.~Hegde, P.-A. Tsai, S.~Huang, V.~Chandra, A.~Parashar, and C.~W. Fletcher,
  ``Mind mappings: enabling efficient algorithm-accelerator mapping space
  search,'' in \emph{Proceedings of the International Conference on
  Architectural Support for Programming Languages and Operation Systems
  (ASPLOS)}, 2021.

\bibitem{centaur-isca2020}
G.~{Henry}, P.~{Palangpour}, M.~{Thomson}, J.~S. {Gardner}, B.~{Arden},
  J.~{Donahue}, K.~{Houck}, J.~{Johnson}, K.~{O’Brien}, S.~{Petersen},
  B.~{Seroussi}, and T.~{Walker}, ``High-performance deep-learning coprocessor
  integrated into x86 soc with server-class cpus industrial product,'' in
  \emph{Proceedings of the International Symposium on Computer Architecture
  (ISCA)}, 2020.

\bibitem{autotm-asplos2020}
M.~Hildebrand, J.~Khan, S.~Trika, J.~Lowe-Power, and V.~Akella, ``Autotm:
  Automatic tensor movement in heterogeneous memory systems using integer
  linear programming,'' in \emph{Proceedings of the International Conference on
  Architectural Support for Programming Languages and Operation Systems
  (ASPLOS)}, 2020.

\bibitem{jia2019beyond}
Z.~Jia, M.~Zaharia, and A.~Aiken, ``Beyond data and model parallelism for deep
  neural networks.'' in \emph{Proceedings of Machine Learning and Systems
  (MLSys)}, 2019.

\bibitem{tpu-isca2016}
N.~P. Jouppi, C.~Young, N.~Patil, D.~Patterson, G.~Agrawal, R.~Bajwa, S.~Bates,
  S.~Bhatia, N.~Boden, A.~Borchers, R.~Boyle, P.~luc Cantin, C.~Chao, C.~Clark,
  J.~Coriell, M.~Daley, M.~Dau, J.~Dean, B.~Gelb, T.~V. Ghaemmaghami,
  R.~Gottipati, W.~Gulland, R.~Hagmann, C.~R. Ho, D.~Hogberg, J.~Hu, R.~Hundt,
  D.~Hurt, J.~Ibarz, A.~Jaffey, A.~Jaworski, A.~Kaplan, H.~Khaitan,
  D.~Killebrew, A.~Koch, N.~Kumar, S.~Lacy, J.~Laudon, J.~Law, D.~Le, C.~Leary,
  Z.~Liu, K.~Lucke, A.~Lundin, G.~MacKean, A.~Maggiore, M.~Mahony, K.~Miller,
  R.~Nagarajan, R.~Narayanaswami, R.~Ni, K.~Nix, T.~Norrie, M.~Omernick,
  N.~Penukonda, A.~Phelps, J.~Ross, M.~Ross, A.~Salek, E.~Samadiani, C.~Severn,
  G.~Sizikov, M.~Snelham, J.~Souter, D.~Steinberg, A.~Swing, M.~Tan,
  G.~Thorson, B.~Tian, H.~Toma, E.~Tuttle, V.~Vasudevan, R.~Walter, W.~Wang,
  E.~Wilcox, and D.~H. Yoon, ``{In-Datacenter Performance Analysis of a Tensor
  Processing Unit},'' in \emph{Proceedings of the International Symposium on
  Computer Architecture (ISCA)}, 2017.

\bibitem{gamma-iccad2020}
S.-C. Kao and T.~Krishna, ``{GAMMA: Automating the HW Mapping of DNN Models on
  Accelerators via Genetic Algorithm},'' in \emph{Proceedings of the
  International Conference on Computer-Aided Design (ICCAD)}, 2020.

\bibitem{khailany2018modular}
B.~Khailany, E.~Krimer, R.~Venkatesan, J.~Clemons, J.~S. Emer, M.~Fojtik,
  A.~Klinefelter, M.~Pellauer, N.~Pinckney, Y.~S. Shao, S.~Srinath, C.~Torng,
  S.~L. Xi, Y.~Zhang, and B.~Zimmer, ``Invited: A modular digital vlsi flow for
  high-productivity soc design,'' in \emph{Design Automation Conference (DAC)},
  2018.

\bibitem{kong2013polyhedral}
M.~Kong, R.~Veras, K.~Stock, F.~Franchetti, L.-N. Pouchet, and P.~Sadayappan,
  ``When polyhedral transformations meet simd code generation,'' in
  \emph{Proceedings of the ACM SIGPLAN Conference on Programming Language
  Design and Implementation (PLDI)}, 2013.

\bibitem{alexnet}
A.~Krizhevsky, I.~Sutskever, and G.~E. Hinton, ``{Imagenet Classification with
  Deep Convolutional Neural Networks},'' in \emph{Proceedings of the Conference
  on Neural Information Processing Systems (NeurIPS)}, 2012.

\bibitem{maeri-asplos2018}
H.~Kwon, A.~Samajdar, and T.~Krishna, ``{MAERI: Enabling Flexible Dataflow
  Mapping over {DNN} Accelerators via Programmable Interconnects},'' in
  \emph{Proceedings of the International Conference on Architectural Support
  for Programming Languages and Operation Systems (ASPLOS)}, 2018.

\bibitem{mullapudi2016automatically}
R.~T. Mullapudi, A.~Adams, D.~Sharlet, J.~Ragan-Kelley, and K.~Fatahalian,
  ``Automatically scheduling halide image processing pipelines,'' \emph{ACM
  Transactions on Graphics (TOG)}, 2016.

\bibitem{naumov2019dlrm}
M.~Naumov, D.~Mudigere, H.-J.~M. Shi, J.~Huang, N.~Sundaraman, J.~Park,
  X.~Wang, U.~Gupta, C.-J. Wu, A.~G. Azzolini, D.~Dzhulgakov, A.~Mallevich,
  I.~Cherniavskii, Y.~Lu, R.~Krishnamoorthi, A.~Yu, V.~Kondratenko, S.~Pereira,
  X.~Chen, W.~Chen, V.~Rao, B.~Jia, L.~Xiong, and M.~Smelyanskiy, ``Deep
  learning recommendation model for personalization and recommendation
  systems,'' 2019.

\bibitem{nowatzki2018hybrid}
T.~Nowatzki, N.~Ardalani, K.~Sankaralingam, and J.~Weng, ``Hybrid
  optimization/heuristic instruction scheduling for programmable accelerator
  codesign,'' in \emph{Proceedings of the International Conference on Parallel
  Architectures and Compilation Techniques (PACT)}, 2018.

\bibitem{nowatzki2013general}
T.~Nowatzki, M.~Sartin-Tarm, L.~De~Carli, K.~Sankaralingam, C.~Estan, and
  B.~Robatmili, ``A general constraint-centric scheduling framework for spatial
  architectures,'' \emph{ACM SIGPLAN Notices}, 2013.

\bibitem{timeloop2019-ispass}
A.~Parashar, P.~Raina, Y.~S. Shao, Y.-H. Chen, V.~A. Ying, A.~Mukkara,
  R.~Venkatesan, B.~Khailany, S.~W. Keckler, and J.~Emer, ``{Timeloop: A
  Systematic Approach to DNN Accelerator Evaluation},'' in \emph{Proceedings of
  the International Symposium on Performance Analysis of Systems and Software
  (ISPASS)}, 2019.

\bibitem{scnn}
A.~Parashar, M.~Rhu, A.~Mukkara, A.~Puglielli, R.~Venkatesan, B.~Khailany,
  J.~Emer, S.~W. Keckler, and W.~J. Dally, ``{SCNN: An Accelerator for
  Compressed-sparse Convolutional Neural Networks},'' in \emph{Proceedings of
  the International Symposium on Computer Architecture (ISCA)}, 2017.

\bibitem{park2013predictive}
E.~Park, J.~Cavazos, L.-N. Pouchet, C.~Bastoul, A.~Cohen, and P.~Sadayappan,
  ``Predictive modeling in a polyhedral optimization space,''
  \emph{International journal of parallel programming}, 2013.

\bibitem{swizzle-asplos2019}
P.~M. Phothilimthana, A.~S. Elliott, A.~Wang, A.~Jangda, B.~Hagedorn,
  H.~Barthels, S.~J. Kaufman, V.~Grover, E.~Torlak, and R.~Bodik, ``Swizzle
  inventor: Data movement synthesis for gpu kernels,'' in \emph{Proceedings of
  the International Conference on Architectural Support for Programming
  Languages and Operation Systems (ASPLOS)}, 2019.

\bibitem{phothilimthana2014chlorophyll}
P.~M. Phothilimthana, T.~Jelvis, R.~Shah, N.~Totla, S.~Chasins, and R.~Bodik,
  ``Chlorophyll: Synthesis-aided compiler for low-power spatial
  architectures,'' in \emph{Proceedings of the ACM SIGPLAN Conference on
  Programming Language Design and Implementation (PLDI)}, 2014.

\bibitem{plasticine-isca2017}
R.~Prabhakar, Y.~Zhang, D.~Koeplinger, M.~Feldman, T.~Zhao, S.~Hadjis,
  A.~Pedram, C.~Kozyrakis, and K.~Olukotun, ``Plasticine: A reconfigurable
  architecture for parallel paterns,'' 2017.

\bibitem{sigma-hpca2020}
E.~{Qin}, A.~{Samajdar}, H.~{Kwon}, V.~{Nadella}, S.~{Srinivasan}, D.~{Das},
  B.~{Kaul}, and T.~{Krishna}, ``Sigma: A sparse and irregular gemm accelerator
  with flexible interconnects for dnn training,'' in \emph{Proceedings of the
  International Symposium on High-Performance Computer Architecture (HPCA)},
  2020.

\bibitem{ragan2013halide}
J.~Ragan-Kelley, C.~Barnes, A.~Adams, S.~Paris, F.~Durand, and S.~Amarasinghe,
  ``Halide: a language and compiler for optimizing parallelism, locality, and
  recomputation in image processing pipelines,'' \emph{Acm Sigplan Notices},
  2013.

\bibitem{yolo}
J.~Redmon, S.~Divvala, R.~Girshick, and A.~Farhadi, ``{You Only Look Once:
  Unified, Real-time Object Detection},'' in \emph{Proceedings of the
  Conference on Computer Vision and Pattern Recognition (CVPR)}, 2016.

\bibitem{rosenfeld2011dramsim2}
P.~Rosenfeld, E.~Cooper-Balis, and B.~Jacob, ``Dramsim2: A cycle accurate
  memory system simulator,'' \emph{IEEE computer architecture letters}, 2011.

\bibitem{shao2019simba}
Y.~S. Shao, J.~Clemons, R.~Venkatesan, B.~Zimmer, M.~Fojtik, N.~Jiang,
  B.~Keller, A.~Klinefelter, N.~Pinckney, P.~Raina, S.~G.~Tell, Y.~Zhang,
  W.~J.~Dally, J.~Emer, C.~T. Gray, B.~Khailany, and S.~W.~Keckler, ``Simba:
  Scaling deep-learning inference with multi-chip-module-based architecture,''
  in \emph{Proceedings of the International Symposium on Microarchitecture
  (MICRO)}, 2019.

\bibitem{shao2019-micro}
Y.~S. Shao, J.~Clemons, R.~Venkatesan, B.~Zimmer, M.~Fojtik, N.~Jiang,
  B.~Keller, A.~Klinefelter, N.~Pinckney, P.~Raina, S.~G. Tell, Y.~Zhang, W.~J.
  Dally, J.~Emer, C.~T. Gray, B.~Khailany, and S.~W. Keckler, ``Simba: Scaling
  deep-learning inference with multi-chip-module-based architecture,'' in
  \emph{Proceedings of the International Symposium on Microarchitecture
  (MICRO)}, 2019.

\bibitem{nvdla-hotchips}
F.~Sijstermans, ``{The NVIDIA Deep Learning Accelerator},'' in \emph{Hot
  Chips}, 2018.

\bibitem{sketching-pldi2005}
A.~Solar-Lezama, R.~Rabbah, R.~Bod\'{\i}k, and K.~Ebcio\u{g}lu, ``Programming
  by sketching for bit-streaming programs,'' in \emph{Proceedings of the ACM
  SIGPLAN Conference on Programming Language Design and Implementation (PLDI)},
  2005.

\bibitem{samsung}
J.~Song, Y.~Cho, J.-S. Park, J.-W. Jang, S.~Leev, J.-H. Song, J.-G. Lee, and
  I.~Kang, ``An 11.5 tops/w 1024-mac butterfly structure dual-core
  sparsity-aware neural processing unit in 8nm flagship mobile soc,'' in
  \emph{Proceedings of the International Solid State Circuits Conference
  (ISSCC)}, 2019.

\bibitem{sutskever2014sequence}
I.~Sutskever, O.~Vinyals, and Q.~V. Le, ``Sequence to sequence learning with
  neural networks,'' in \emph{Proceedings of the Conference on Neural
  Information Processing Systems (NeurIPS)}, 2014.

\bibitem{tillet2019triton}
P.~Tillet, H.~Kung, and D.~Cox, ``Triton: an intermediate language and compiler
  for tiled neural network computations,'' in \emph{Proceedings of the 3rd ACM
  SIGPLAN International Workshop on Machine Learning and Programming
  Languages}, 2019.

\bibitem{wavenet}
A.~van~den Oord, S.~Dieleman, H.~Zen, K.~Simonyan, O.~Vinyals, A.~Graves,
  N.~Kalchbrenner, A.~Senior, and K.~Kavukcuoglu, ``Wavenet: A generative model
  for raw audio,'' 2016.

\bibitem{vasilache2018tensor}
N.~Vasilache, O.~Zinenko, T.~Theodoridis, P.~Goyal, Z.~DeVito, W.~S. Moses,
  S.~Verdoolaege, A.~Adams, and A.~Cohen, ``Tensor comprehensions:
  Framework-agnostic high-performance machine learning abstractions,'' 2018.

\bibitem{transformer}
A.~Vaswani, N.~Shazeer, N.~Parmar, J.~Uszkoreit, L.~Jones, A.~N. Gomez, L.~u.
  Kaiser, and I.~Polosukhin, ``Attention is all you need,'' in
  \emph{Proceedings of the Conference on Neural Information Processing Systems
  (NeurIPS)}, 2017.

\bibitem{scaledeep}
S.~Venkataramani, A.~Ranjan, S.~Banerjee, D.~Das, S.~Avancha, A.~Jagannathan,
  A.~Durg, D.~Nagaraj, B.~Kaul, P.~Dubey, and A.~Raghunathan, ``{ScaleDeep: A
  Scalable Compute Architecture for Learning and Evaluating Deep Networks},''
  in \emph{Proceedings of the International Symposium on Computer Architecture
  (ISCA)}, 2017.

\bibitem{Xie2016resnext}
S.~Xie, R.~Girshick, P.~Dollar, Z.~Tu, and K.~He, ``Aggregated residual
  transformations for deep neural networks,'' in \emph{Proceedings of the IEEE
  Conference on Computer Vision and Pattern Recognition (CVPR)}, 2017.

\bibitem{interstellar-asplos2020}
X.~Yang, M.~Gao, Q.~Liu, J.~Setter, J.~Pu, A.~Nayak, S.~Bell, K.~Cao, H.~Ha,
  P.~Raina, C.~Kozyrakis, and M.~Horowitz, ``Interstellar: Using halide’s
  scheduling language to analyze dnn accelerators,'' in \emph{Proceedings of
  the International Conference on Architectural Support for Programming
  Languages and Operation Systems (ASPLOS)}, 2020.

\bibitem{cambricon}
S.~Zhang, Z.~Du, L.~Zhang, H.~Lan, S.~Liu, L.~Li, Q.~Guo, T.~Chen, and Y.~Chen,
  ``{Cambricon-X: An Accelerator for Sparse Neural Networks},'' in
  \emph{Proceedings of the International Symposium on Microarchitecture
  (MICRO)}, 2016.

\bibitem{janus-cgo2019}
R.~{Zhou} and T.~M. {Jones}, ``Janus: Statically-driven and profile-guided
  automatic dynamic binary parallelisation,'' in \emph{International Symposium
  on Code Generation and Optimization (CGO)}, 2019.

\end{thebibliography}
